\documentclass[10pt]{article}

\usepackage[
  letterpaper,
  left=1in,
  right=1in,
  top=1in,
  bottom=1in
]{geometry}

\usepackage{setspace}
\usepackage[T1]{fontenc}
\usepackage{lmodern}      
\usepackage{microtype}
\usepackage{tabularx}
\usepackage{graphicx}
\usepackage{xcolor}
\usepackage{booktabs}  
\usepackage{multirow}  

\usepackage{titlesec}

\titleformat{\section}
  {\normalfont\large\bfseries}{\thesection}{1em}{}
\titleformat{\subsection}
  {\normalfont\normalsize\bfseries}{\thesubsection}{1em}{}
\titleformat{\subsubsection}
  {\normalfont\normalsize\itshape}{\thesubsubsection}{1em}{}

\titlespacing*{\section}{0pt}{2.5ex plus 1ex minus .2ex}{1.5ex}
\titlespacing*{\subsection}{0pt}{2ex plus 0.8ex minus .2ex}{1ex}

\usepackage[style=numeric-comp, sorting=none, backend=biber]{biblatex}
\usepackage{glossaries}
\usepackage{multirow}
\usepackage{xcolor}
\usepackage{tabularx}

\newacronym{ar}{AR}{augmented reality}
\newacronym{aar}{AAR}{audio augmented reality}
\newacronym{vr}{VR}{virtual reality}
\newacronym{mr}{MR}{mixed reality}
\newacronym{hri}{HRI}{human-robot interaction}
\newacronym{rosas}{RoSAS}{Robot Social Attribute Scale}
\newacronym{hmd}{HMDs}{head-mounted displays}
\newacronym{avas}{AVAS}{acoustic vehicle alerting systems}
\newacronym{art}{ART}{Aligned Rank Transform}

\begin{document}

\title{The Role of Consequential and Functional Sound in Human–Robot Interaction: Toward Audio Augmented Reality Interfaces}

\author{
Aliyah Smith \\
Stanford University \\
\texttt{aliyah1@stanford.edu}
\and
Monroe Kennedy III \\
Stanford University \\
\texttt{monroek@stanford.edu}
}
\date{}

\maketitle
\glsdisablehyper


\begin{abstract}
Robot sound—encompassing both consequential operational noise and intentionally designed auditory cues—plays an important role in human–robot interaction (HRI). Developing a deeper understanding of how robot sounds influence human experience, and how technologies such as augmented reality (AR) modulate these effects, can enable the design of more socially acceptable robots and more effective, intuitive human–robot interfaces. In this work, we present a three-part mixed-methods study (N = 51) that investigates (i) the effects of consequential robot sounds on human perception under varying degrees of physical colocation, (ii) human accuracy in localizing spatial audio cues delivered via augmented reality, and (iii) the use of augmented spatial audio cues for functional and transformative communication during collaborative handover tasks, in comparison to non-AR sound designs. Contrary to prior findings, our results indicate that the consequential sounds of a Kinova Gen3 manipulator did not negatively affect participants’ perceptions of the robot. Participants demonstrated high accuracy in localizing lateral spatial cues, whereas frontal cues proved more challenging, delineating conditions under which spatial auditory feedback is most effective. Qualitative findings further reveal that augmented spatial audio cues can simultaneously convey task-relevant information while fostering a sense of warmth and reducing user discomfort during interaction. Together, these findings elucidate the perceptual effects of consequential robot sound and position sound—particularly augmented spatial audio—as a meaningful yet underutilized design resource for human–robot interaction.
\end{abstract}



\section*{Keywords}{Human-robot interaction, audio augmented reality, consequential sound, functional sound}



\section{Introduction}
With the increasing presence of embodied agents in public spaces, workplaces, and homes, understanding how to foster positive human perceptions of robots is of paramount importance. A key driver of these perceptions is robots’ use of nonverbal cues, which strongly shape human–robot interaction by influencing users’ emotional experiences and behavior \cite{urakami}. Among these cues, robot-generated sounds play a particularly salient role and are commonly categorized into four primary types: consequential, functional, transformative, and emotional \cite{zhang}. Consequential sounds emerge naturally from a robot’s mechanical operation and are inherently tied to the physical characteristics of a given platform \cite{zhang, allen}. Functional sounds are intentionally designed to convey information, such as signaling a robot’s state, actions, or intentions \cite{nwagwu2024benefits}. Transformative sounds modify or mask a robot’s inherent acoustic profile to shape its perceived identity, presence, or approachability \cite{rogel2025sounds}. Emotional sounds explicitly express affective states or social intent, supporting empathy and interpersonal connection \cite{latupeirissa2023pepperosc}. Together, these auditory cues strongly influence how humans perceive, interpret, and engage with robots, underscoring sound’s importance as a core interaction modality in \gls{hri}. Despite this significance, empirical investigations of robot sound remain constrained. In particular, many sound-focused studies—especially those examining the effects of consequential sound—have relied on online experiments or specific robotic platforms due to the logistical challenges of conducting in-person studies. As a result, there remains a limited understanding of how robot-generated sounds from robotic manipulators shape human perception and experience in collocated interaction settings.

Augmented reality has become an increasingly powerful medium for interactive interface design with the widespread adoption of wearable \gls{hmd} \cite{azuma_ar}. Within the broader landscape of \gls{ar} research, however, \gls{aar} remains a comparatively underexplored domain \cite{Yang2022AudioAR}. While recent advances in \gls{ar} technologies have driven substantial progress in visual augmentation—enabling virtual objects to be seamlessly integrated into the physical environment—research on auditory augmentation has lagged behind. In \gls{aar}, virtual auditory cues are spatially and contextually embedded within a user’s real-world soundscape \cite{almeida}, creating unique opportunities to enhance perception \cite{Chang28042025}, situational awareness \cite{shayman2024addition}, and interaction \cite{broderick}. Despite this promise, \gls{aar} has received little systematic attention within the \gls{hri} community. This gap is particularly striking given the growing adoption of \gls{ar} headsets and interface design, the central role of sound in human–robot interaction, and the increasing prevalence of collocated, embodied robotic systems. As a result, there remains limited understanding of how augmented and spatialized auditory cues can be leveraged for both functional and transformative purposes—communicating task-relevant information to humans while also shaping their perceptions, expectations, and experiences of robots.

To address these gaps, this work investigates the intersection of sound, \gls{aar}, and \gls{hri}, with the goal of understanding how both naturally occurring consequential sounds produced by a Kinova Gen3 manipulator and intentionally designed functional auditory cues shape human perception and interaction. We present a three-part, in-person study. First, we replicate and extend prior work on the effects of consequential robot sound on human perception, examining how varying collocated settings and sound conditions influence user responses. Second, we evaluate humans’ ability to detect and localize augmented spatial audio cues to establish their perceptual viability and motivate the use of \gls{aar} in \gls{hri}. Third, we compare three sound conditions in a collaborative human–robot task, investigating how each design—including a specific \gls{aar}-based approach—affects communication, spatial awareness, and perceptions of robot behavior.

By bridging perspectives from auditory augmentation and \gls{hri} sound design, this work contributes empirical evidence and design insights toward a deeper understanding of how robot sound influences human experience. More broadly, it positions sound as an expressive, informative, and underutilized interaction channel for shaping effective and engaging human–robot collaboration.

The key contributions of this paper are as follows:
\begin{itemize}
    \item A rigorous in-person replication of prior work (N = 48) examining the effects of consequential robot sounds on human perception \cite{allen}, extending earlier findings beyond online settings to a collocated interaction with a physical robotic manipulator.
    \item An empirical characterization of human spatial sound localization within a constrained robotic workspace (N = 51), providing foundational evidence for the feasibility and limitations of spatial auditory cues (\gls{aar}) as an interaction modality in \gls{hri}.
    \item A controlled comparative study (N = 41) evaluating consequential, functional, and spatial (\gls{aar}) auditory feedback during a human–robot collaborative task, demonstrating how different sound design strategies differentially influence user perception and experience (i.e., how these designs affect feelings of robot warmth and competence, and feelings of discomfort).
    \item Actionable design insights and guidelines for sound-based interaction in \gls{hri}, synthesizing quantitative and qualitative findings to inform the intentional use of functional and spatial (\gls{aar}) sound in future human–robot interfaces.
\end{itemize}


\section{Related Work}

Nonverbal sound, while still underexplored, has gained increasing attention in the robotics community over recent years \cite{zhang}, with research exploring how auditory cues can influence human perception, interaction, and task performance. Prior work can be broadly categorized into consequential, functional, transformative, and emotional sounds, each highlighting distinct mechanisms through which sound shapes human-robot interaction. In this work, we focus on consequential, functional, and transformative sounds. 

\subsection{Studies on Consequential, Functional, and Transformative Sounds in HRI}

Research on consequential sounds—those naturally produced by robots—has primarily examined their effects on localization, perception, and detectability. Wessels et al. showed that localization accuracy and precision varied by robot type due to their inherent sounds, and found that participants preferred the sounds linked to poorer performance \cite{wessels2025auditorylocalizationassessmentconsequential}. Allen et al. found that high-pitched or loud robot sounds elicited negative impressions, while rhythmic, natural, or functional sounds enhanced predictability and user comfort \cite{allen, savery2023sound}, echoing findings from \cite{wang, trovato, tennent, moore, jouaiti}. \textit{Collectively, these studies highlight how consequential sounds shape users’ perception and awareness of robots; however, most have been conducted in online settings. Moreover, such sounds are inherently tied to specific robotic platforms, thus further in-person investigations with additional robot types are needed to generalize these findings.}

Research on functional sounds—auditory cues intentionally designed to convey information—remains limited. Orthmann et al. showed that nonverbal sounds can communicate robot attributes such as size, speed, and availability, though comprehension declines with overlapping sounds \cite{orthmann}. Zahray et al. examined gesture sonification to express robot movement with mixed perceptual outcomes \cite{zahray_gesture}, while Okimoto et al. found that auditory cues for task completion improved collaboration fluency \cite{okimoto}. Boos et al. demonstrated that nonverbal auditory and visual cues can convey critical information in military contexts, though ambiguity remains a challenge \cite{boos2019conveying}. Similarly, Cha et al. found that designed robot sounds enhance auditory localization \cite{cha}. \textit{Collectively, these works underscore the communicative and perceptual value of functional sounds but remain limited by narrow task contexts, small-scale evaluations, and a lack of systematic investigation into which information is most effectively conveyed through sound.}

Studies on transformative sounds—auditory designs that intentionally alter perception or convey affect—span domains such as safety, entertainment, and social interaction. Rogel et al. showed that musical attributes like tempo, pitch, and timbre influence perceived urgency and safety \cite{rogel}, while Cuan et al. demonstrated that music-based motion-sound mappings enhance engagement and robot perception \cite{cuan2}. Sound has also been shown to shape impressions of robots: Latupeirissa et al. and Robinson et al. found that movement sonification affects perceived personality and fluidity \cite{latupeirissa, robinson_smooth_operator}, with similar effects observed for soft and humanoid robots by Frederiksen et al. and Bresin et al. \cite{frederiksen, bresin2021robust}. Nwagwu et al. reported that transformative character-like sounds increase warmth, competence, and localizability in collaboration \cite{nwagwu2024benefits}, and Zhang et al. advanced the design and study of transformative robot sounds through qualitative and perceptual analyses \cite{zhang_kawaii}. \textit{These studies demonstrate the potential of transformative sounds to shape human perceptions of robots; however, few have examined how such sounds can also serve functional purposes by intentionally conveying specific information beyond gesture sonification.}

\subsection{Spatial Audio and Audio Augmented Reality}

Spatial audio remains largely underexplored within \gls{hri}. To date, only one known study by Robinson et al. has examined spatially distributed sound in a human–robot context, using external speakers in conjunction with a robot \cite{robinson}. While their findings indicate that spatialized sound can influence participant behavior and perception, the study did not include non-spatial comparison conditions or structured task-based evaluations, limiting conclusions about the functional benefits of spatial audio in \gls{hri}.

In contrast, prior work in \gls{vr} and \gls{ar} has demonstrated that spatial audio can enhance immersion, perceptual clarity, and task performance \cite{Chang28042025, jenny}. For example, Chang et al. reported improved sound recall and intelligibility for lateral audio sources when compared to traditional surround sound systems \cite{Chang28042025}. \textit{Drawing on these findings, integrating spatial audio and \gls{aar} into \gls{hri} holds promise for unifying functional and transformative auditory cues, thereby enabling richer and more intuitive human–robot interactions.}

\subsection{Design of Functional and Transformative Sounds}
Functional sound, outside the context of robotics, is ubiquitous in everyday life, despite receiving relatively limited formal study in the literature. Many common devices—such as microwaves, mobile phones, and automobiles—incorporate auditory cues to communicate information to users (e.g., signaling task completion, notifications, or safety). These sounds convey functionally relevant information that supports user awareness and decision-making. Certain auditory cues, such as the sound of a car engine or an \gls{avas} \cite{muller, emerson, wessels}, can be considered functional despite also being categorized as consequential or transformative, highlighting the overlap between these sound classes.

Drawing inspiration from such everyday examples, we ground the design of two sound conditions for a collaborative human–robot task. First, we emulate mobile phone–style notification sounds to create a discrete, purely functional auditory cue emitted from an external speaker. Second, we emulate the continuous sound profile of an \gls{avas} to create a combined functional and transformative auditory cue delivered via an \gls{ar} headset. Further details regarding the sound designs and their implementation are provided in Section 3.

\subsection{Experimental Design and Motivation}
The experimental design of this three-part study builds upon several prior works. Experiment A is informed by Allen et al. \cite{allen}, who found that consequential robot sounds were associated with more negative perceptions across five different robots. While their study provided a comprehensive between-subjects analysis of consequential sound, it presents two key limitations. First, the study did not include high degree-of-freedom robotic manipulators, which can appear or sound more intimidating to users, thereby limiting the generalizability of the findings. Although Allen et al. hypothesized that different robots would elicit varying levels of negative perception, their analysis did not include manipulators, leaving this hypothesis untested. Second, all sounds were presented remotely via video, constraining insight into collocated human–robot interactions. To address these limitations, we replicate and extend Allen et al.’s work in an in-person setting using a robotic manipulator. Specifically, we compare participants who are physically collocated with the robot to those who observe the robot via video, under both full-sound and sound-reduced conditions. Izui et al. conducted a similar study that tested the correlation between participant impressions of avatars, real, sound-proof, and recorded humanoids \cite{izui2020correlation}. However, their study only partially focused on the effects of motor noise, and the avatar, recorded, and soundproof conditions were all silent. We hypothesize that participants exposed to the consequential sounds of the Kinova Gen3 will report more negative perceptions than those interacting under silent conditions, consistent with prior findings. Additionally, we hypothesize that collocated participants will report more negative perceptions than those interacting via video. This design enables us to assess whether prior online findings generalize to in-person contexts and provides further insight into the perceptual effects of consequential robot sounds in realistic \gls{hri} settings.

Experiment B builds on prior work in spatial sound localization \cite{carlini, yost, kolarik2016auditory, getzmann2007localization, stevens}, which has examined humans’ ability to localize stationary and moving sounds presented via speakers arranged in three-dimensional space. Much of this research has been conducted in specialized laboratory environments under tightly controlled conditions (e.g., quiet, dark settings), limiting its applicability to everyday interaction contexts. In contrast, the \gls{hri} community lacks empirical insight into the effectiveness of \gls{ar} headsets as a medium for spatial sound delivery, as well as users’ ability to accurately identify sound sources in less controlled, real-world environments. Drawing on prior findings in auditory perception, we hypothesize that participants will localize lateral spatial sounds (left and right) with high accuracy, but will exhibit reduced accuracy for sounds originating from frontal directions. Characterizing these performance differences is critical for evaluating the perceptual viability and practical utility of \gls{aar} in \gls{hri} contexts. Accordingly, our work addresses this gap by assessing how accurately humans can localize spatial sounds delivered through \gls{ar} \gls{hmd}, thereby motivating the broader adoption of augmented auditory cues in collaborative human–robot interactions.

Experiment C draws inspiration from Nwagwu et al. \cite{nwagwu2024benefits}, who found that character-like sounds elicited warmer social perceptions of a robot teammate. In their within-subjects study, participants completed a joint task with a social robot under three sound conditions—consequential, functional, and character-like—where the character-like condition combined transformative and emotional sounds. Similarly, Experiment C investigates the combination of different sound types using \gls{aar}, specifically integrating transformative and functional sounds. This approach is particularly appropriate for a non-social robot, which is not intended to convey emotions directly. The \gls{hri} community could benefit from additional insights into novel augmented sound designs that merge multiple robot sound types, as well as guidance on the role of functional sounds, including which types of information are most effectively communicated. Building on prior work, we compare a consequential sound condition with two custom designs: a non-AR functional sound and a spatially augmented design delivered via \gls{aar} (inspiration for these designs is discussed in Section 2.3). We hypothesize that both the functional and spatial sounds will enhance participants’ perceptions of the robot as warmer and more competent, while also reducing feelings of discomfort. Additionally, we hypothesize that participants will be able to discern the specific functional information conveyed by the two sound designs, despite not being explicitly informed of their purpose or existence. Finally, to gather broader insights, we administered a structured survey at the end of the study to assess participants’ opinions on functional sound in \gls{hri}. Collectively, these findings contribute to the growing body of knowledge on sound design in \gls{hri}, particularly regarding the integration of multiple sound types to enhance communication, spatial awareness, and overall user experience. Further details on the specific experimental design elements for Experiments A, B, and C are provided in Section 3.

\section{Methodology}
\begin{figure}
    \centering
    \includegraphics[width=1.0\linewidth]{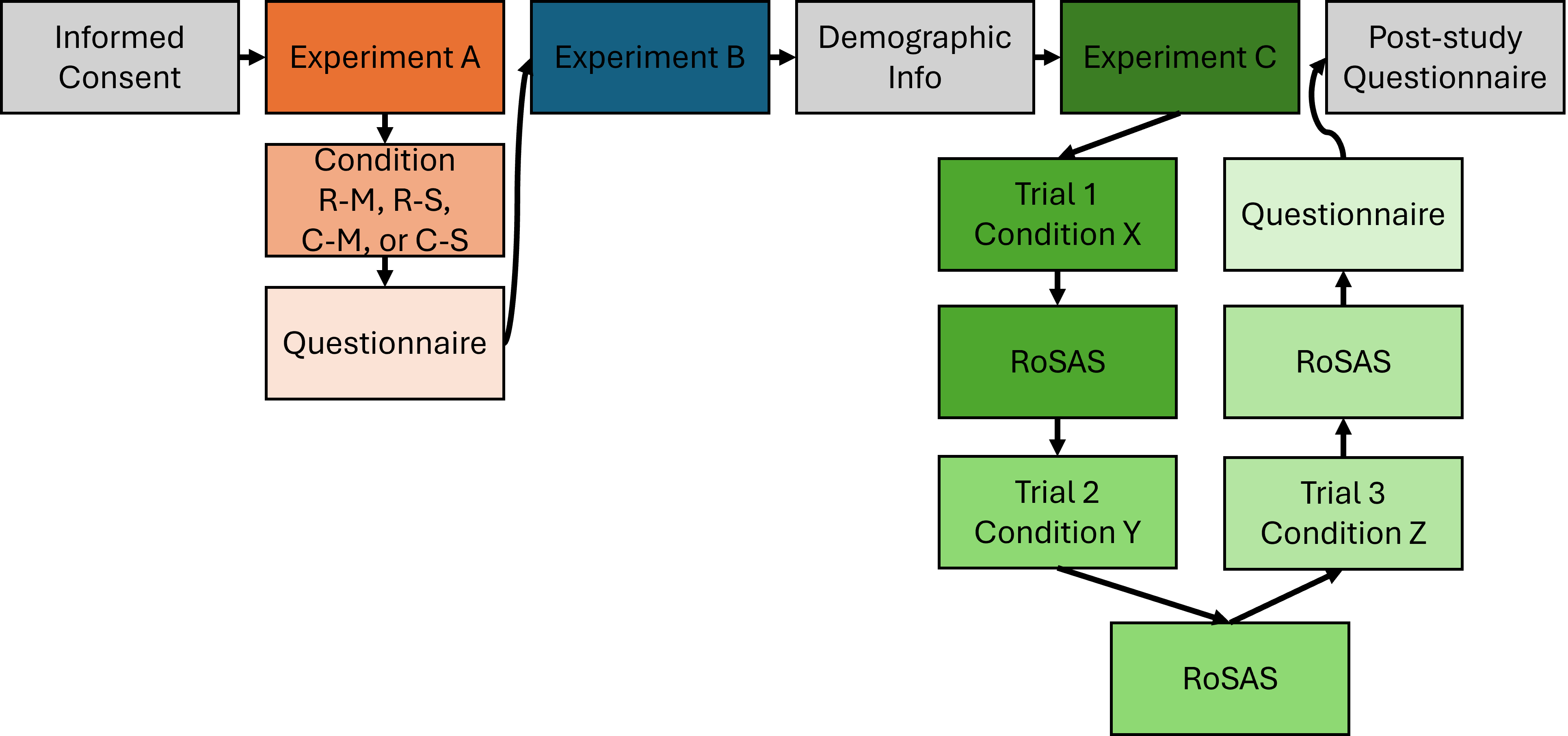}
    \caption{Three-part study procedure.}
    \label{fig:methods}
\end{figure}

We recruited participants for a three-part study investigating consequential, functional, and augmented spatial sounds in the context of \gls{hri} under IRB protocol 65022. Informed consent was obtained at the beginning of each session, demographic information was gathered at the mid-point while the experimenter prepared the final experiment, and participants completed a short post-study questionnaire at the end of the session, which lasted approximately 45 minutes in total. Figure \ref{fig:methods} outlines the study procedure. Participants were compensated with a \$15 gift card for their time. 

\subsection{\textbf{Experiment A: An In-person Replication of the Study on the Effects of Consequential Sounds on Human Perception of Robots}}
This experiment aims to study the effects of consequential sounds on human perceptions of the Kinova Gen3 manipulator. 

\subsubsection{Hypotheses} 
Below are the hypotheses for Experiment A, reiterated from Section 2.4.

\indent \textbf{H1}: When observing the robot with sound (either via video or in collocated settings), participants will report more negative affective responses, lower levels of liking, and a reduced desire for physical collocation.

\indent \textbf{H2}: Participants physically collocated with the robot will report more negative perceptions of the robot than participants exposed to consequential robot sounds via video recordings.

\subsubsection{Experimental Study Design}
\begin{figure}
    \centering
    \includegraphics[width=0.8\linewidth]{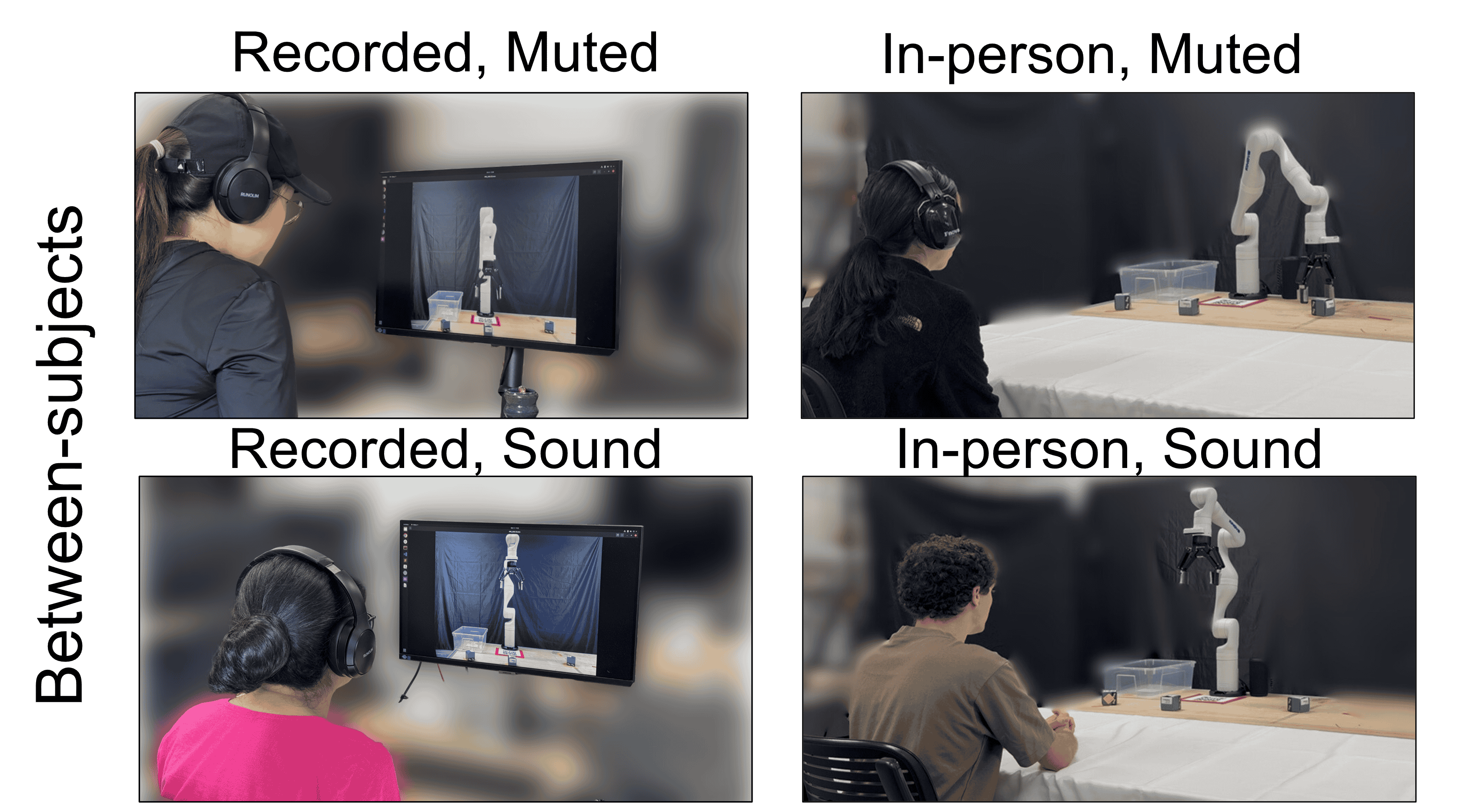}
    \caption{The four experimental conditions for Experiment A are shown. The Kinova Gen3 manipulator was selected for its suitability in collaborative and household tasks (e.g., TidyBot \cite{tidybot}). Participants were assigned to conditions using a quasi-random procedure to ensure equal group sizes.}
    \label{fig:experimentAconditions}
\end{figure}

To establish a perceptual baseline for the effects of sound on user responses to the Kinova Gen3 manipulator, we replicated and extended a prior between-subjects study using video-based stimuli and surveys \cite{allen}. This design enabled us to assess whether consequential robot sounds elicit more negative perceptions and whether these effects differ between collocated and video-based observation. Replication was chosen to support comparability with prior findings while extending the investigation to in-person interaction contexts.

We evaluated four experimental conditions: \textit{Recorded and Muted}, \textit{Recorded with Sound}, \textit{Collocated and Muted}, and \textit{Collocated with Sound}. Figure \ref{fig:experimentAconditions} illustrates the experimental setup for each condition. In the recorded conditions, participants viewed videos of the robot on a monitor, either with or without audible consequential sounds. In the in-person conditions, participants were seated across from the robot and either experienced the robot’s full consequential sound or wore heavy-duty noise-canceling headphones to attenuate auditory input.

Across all conditions, the Kinova Gen3 performed a standardized one-minute pick-and-place task involving representative arm and gripper motions. No task-related context or explanation was provided prior to observation, allowing participants’ perceptions to be shaped primarily by visual and auditory cues. This controlled design isolates the role of consequential sound while enabling direct comparison across observation modalities.

\begin{table}[t]
\centering
\small
\caption{Experiment A: Questionnaire items used to evaluate participant perceptions of the robot. Likert-scale items were rated on a 1--7 scale. Sound-related open-ended questions were presented only to participants in the Sound conditions.}
\label{tab:expA_questionnaire_items}
\begin{tabular}{p{13cm}}
\toprule

\textit{\textbf{Likert-scale items (1--7)}} \\
\midrule

\textit{Associated Affect} \\
How anxious did the robot make you feel? \\
How agitated did the robot make you feel? \\
How unsafe did the robot make you feel? \\
How uncomfortable did the robot make you feel? \\

\textit{Distraction} \\
How distracted were you by the robot? \\

\textit{Collocation} \\
How would you rate your desire for this robot to be collocated with you in your home? \\
How would you rate your desire for this robot to be collocated with you in your workplace? \\
How would you rate your desire for this robot to be collocated with you in a public space? \\

\textit{Like} \\
How much do you like the physical appearance of the robot? \\
How much do you like the movements of the robot? \\
How much do you like the robot overall? \\

\midrule
\textit{\textbf{Open-ended items}} \\
\midrule
Is there anything specific that you liked or disliked about the robot? \\
What features of this robot make you happy or unhappy to be in the same space as this robot? \\
How do you feel about the robot sounds overall? (Sound conditions only) \\
Describe how you would want this robot to sound. (Sound conditions only) \\
If you could change the robot sounds, what changes would you make? (Sound conditions only) \\
Any other comments about the robot? \\

\bottomrule
\end{tabular}
\end{table}

\subsubsection{Measures}
Following the observation period, participants completed an 11-item Likert-scale questionnaire designed to assess the extent to which they were negatively affected or distracted by the robot, as well as their overall liking of the robot and willingness to be collocated with it. Participants also responded to a series of open-ended questions regarding the robot’s overall appearance. Those assigned to the sound conditions were additionally asked questions specific to the robot’s auditory characteristics. Table \ref{tab:expA_questionnaire_items} summarizes the questionnaire; for further details regarding the questionnaire, we refer readers to \cite{allen}.

\subsubsection{Analysis}
Likert questionnaire responses were analyzed using a two-way analysis of variance (ANOVA) to examine the effects of sound condition and observation modality. A significance threshold of $\alpha = 0.05$ was adopted, consistent with established conventions in \gls{hri} and behavioral research, to balance sensitivity to meaningful effects with control of Type I error. The assumption of normality was evaluated using the Shapiro–Wilk test. Although prior work analyzed similar experimental conditions using least-squares regression, we employed a two-way ANOVA to evaluate the effects of sound condition and collocation, as well as their interaction, on participant responses. For factorial experimental designs with categorical predictors, two-way ANOVA is mathematically equivalent to an ordinary least-squares regression model with dummy-coded predictors and interaction terms, differing primarily in presentation rather than underlying inference. We adopt the ANOVA framework here for clarity and consistency with standard reporting practices in \gls{hri} studies.

For the qualitative feedback on the sound-related questions, we conducted a thematic coding analysis to identify common themes across all three questions.

\subsection{\textbf{Experiment B: A Study on Spatial Sound Localization}}
The aim of this experiment is to evaluate participants’ ability to accurately localize static augmented sounds in three-dimensional space.

\subsubsection{Hypotheses} Below is the hypothesis for Experiment B, reiterated from Section 2.4.

\indent \textbf{H3}: Participants will localize static sounds originating from lateral directions (i.e., larger azimuth angles) more accurately than sounds originating from directly in front of them.

\subsubsection{Experimental Study Design}
To assess participants’ ability to discriminate spatial sounds, we conducted an \gls{aar} experiment in which participants were seated across from the robot while wearing a HoloLens~2 \gls{ar} headset that rendered spatialized 360° audio. Participants’ localization accuracy was evaluated across three experimental scenes, each containing virtual markers that denoted the positions of static audio sources.

\begin{figure}
    \centering
    \includegraphics[width=1.0\linewidth]{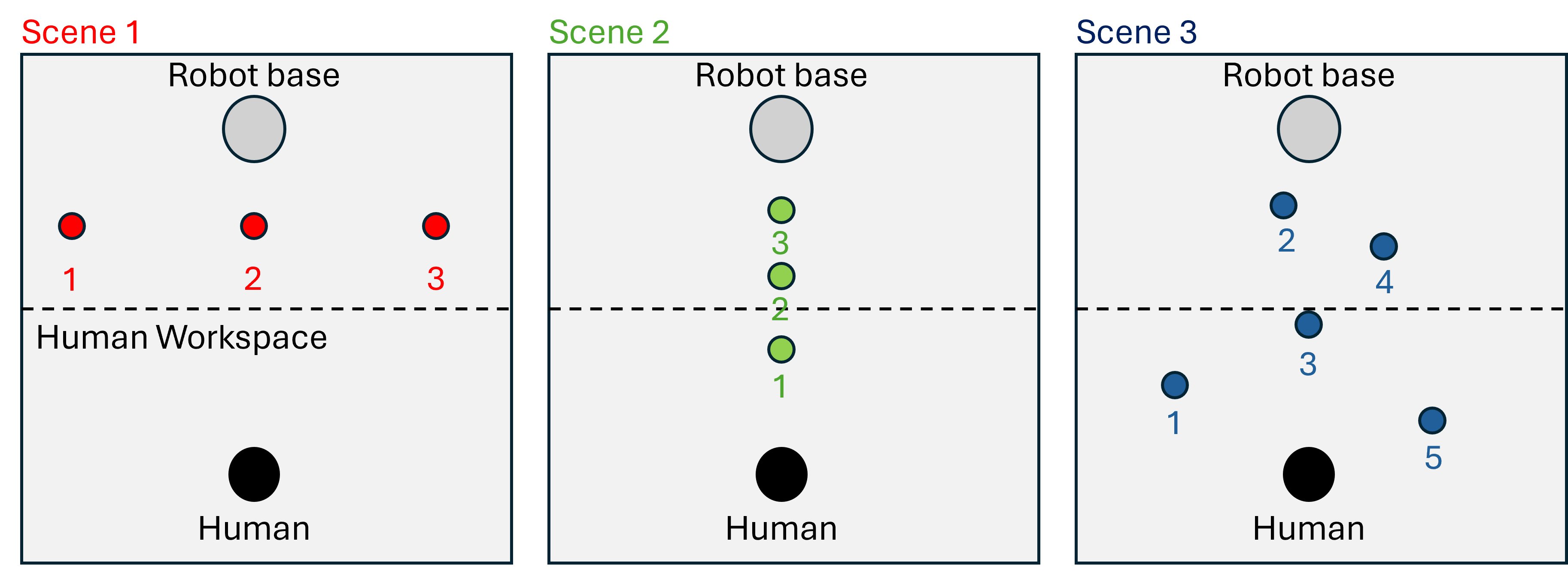}
    \caption{A top-down diagram of the augmented audio source locations within the human-robot workspace for the three scenes in Experiment B. While the robot is not moving in this experiment, the audio sources could represent end-effector positions across the workspace. Scenes 1 and 2 were intentionally designed to test lateral vs frontal cues. Scene 3 was designed to test lateral and frontal cues in tandem.}
    \label{fig:part2_topdown}
\end{figure}

In Scene 1, the application displayed three red spheres distributed across the robot’s workspace. Scenes~2 and~3 presented three green spheres and five blue spheres, respectively, with each sphere corresponding to a distinct three-dimensional audio source. In Scenes~1 and~2, sounds were initially presented sequentially with visual feedback: the corresponding spheres oscillated while the sounds were played to familiarize participants with the audio–spatial mapping. This sequence was played twice. In the subsequent two trials, sounds were presented in a randomized order without visual cues, and participants verbally identified the perceived source of each sound. In Scene~3, no prior visual feedback or familiarization was provided; sounds were presented randomly without sphere oscillations, requiring participants to rely solely on auditory cues for localization. Figure \ref{fig:part2_topdown} shows a top-down diagram of the audio sources in the human-robot workspace and Figure~\ref{fig:experimentBresults} visualizes the three experimental scenes as rendered in \gls{ar}.

\subsubsection{Measures}
For each scene, the true sequence of audio sources and the corresponding participant-identified (predicted) sequence were recorded across both trials.

\subsubsection{Analysis}
Quantitative results were analyzed using confusion matrices that compared the predicted and true labels for each scene across all participants. These matrices were used to compute classification accuracy and examine error patterns, providing insight into overall spatial sound localization performance and potential sources of misclassification.

\subsection{\textbf{Experiment C: Exploring Functional Sounds in Human-Robot Collaboration}}
This experiment examines how two functional audio designs influence human–robot collaboration: one non-\gls{aar} design and one \gls{aar}-based design. Both designs are compared to each other as well as to a non-augmented baseline condition that includes only the robot’s natural consequential audio.

\subsubsection{Functional Sound Designs}
The two functional sound designs were informed by real-world examples and prior research in \gls{hri}, sound design, and auditory localization. The \textit{Functional} (non-spatial) design, developed by the authors, emulates everyday auditory cues from familiar devices such as smartphones and household appliances (see Section 2.3 for additional details on inspiration). Delivered via a Bluetooth speaker positioned near the Kinova manipulator’s base, this design included six distinct sounds corresponding to key robot states: startup, stop, home, retract, target identification, and handover readiness.

The \textit{Spatial} (functional and transformative) design, implemented using \gls{aar}, was inspired by Acoustic Vehicle Alerting Systems (AVAS) in electric vehicles (see Section 2.3 for additional details). Developed in Unity 2020.3 and deployed on the HoloLens 2 \gls{ar} headset, the design used a calming guitar loop \cite{guitarloop} that dynamically moved through three-dimensional space to preemptively signal the robot’s planned gripper position, with sound movements occurring approximately 3s before the robot’s motion, a timing determined empirically to strengthen the perceived connection between auditory cues and robot movement. This approach was chosen to convey robot movement entirely through auditory cues, minimizing reliance on visual feedback that could distract users during the collaborative task. While the \textit{Functional} design explicitly communicates robot states, the \textit{Spatial} design conveys the same information (e.g., startup, stop, handover readiness) implicitly through dynamic spatial audio. Thus, although the two designs differ in presentation, they were intentionally structured to communicate equivalent functional information. 

A fiducial marker aligned reference frames between the robot and headset, with communication managed via ROS\# \cite{ros-sharp}. The spatial mixer applied a logarithmic volume rolloff between 0.1m and 2.0m and amplified lateral movements by 1.5× along the horizontal axis to enhance perceptual salience within the confined workspace. These parameters were determined empirically, guided by prior findings indicating decreased localization accuracy at small azimuth angles. As the gripper approached the participant, volume increased, and the sound ceased upon completion of motion, thereby communicating the robot’s trajectory, status, and safety in real time.

\begin{figure}
    \centering
    \includegraphics[width=1.0\linewidth]{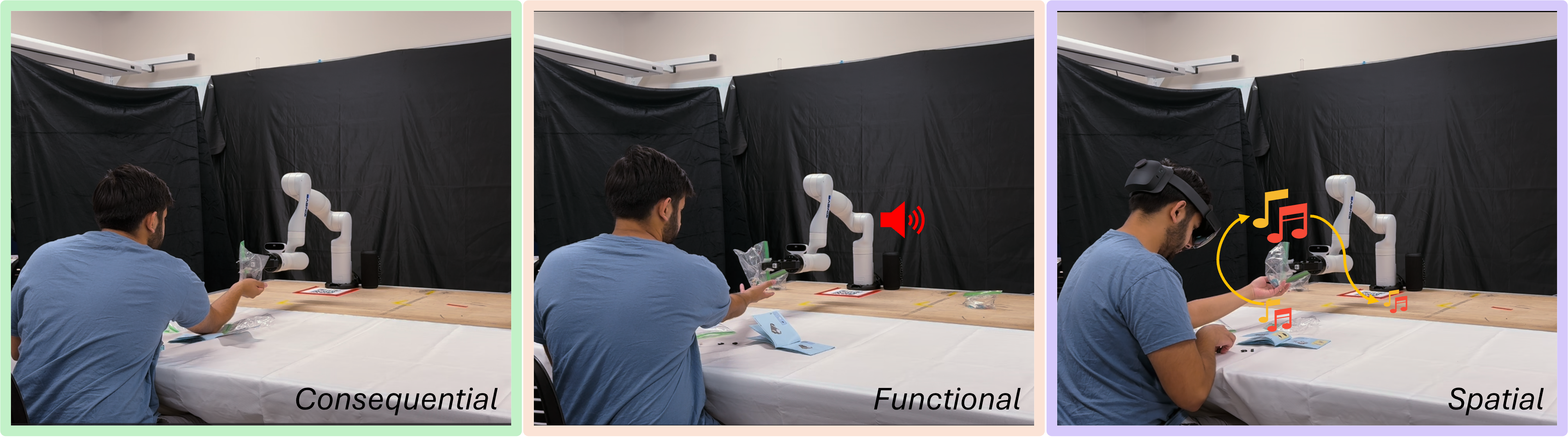}
    \caption{Experiment C involved three sound conditions. Participants completed a LEGO build across three trials, with each trial corresponding to a different sound condition during which the robot handed over additional LEGO pieces. In the \textit{Consequential} condition, participants heard only the robot’s natural sounds. In the \textit{Functional} condition, participants heard augmented sounds (e.g., beeps and chimes) played from a speaker positioned next to the robot. In the \textit{Spatial} condition, participants experienced a continuous musical piece whose source moved according to the robot’s end-effector trajectory, moving ~3 seconds ahead of actual robot movement. These sounds in this condition were delivered using \gls{aar} on the HoloLens 2. The order of conditions was randomized to control for potential order effects.}
    \label{fig:experimentC_sound_conditions}
\end{figure}

\subsubsection{Hypotheses} Below are the hypotheses for Experiment C, reiterated from Section 2.4.

\indent \textbf{H4}: Adding functional sound will increase participants' feelings of robot competence and reduce participants' feelings of discomfort compared to consequential sounds alone.

\indent \textbf{H5}: Adding spatial sound will increase participants' feelings of robot warmth and competence, and reduce participants' feelings of discomfort compared to consequential sounds or purely functional sounds.

\indent \textbf{H6}: Participants will accurately interpret the intended meaning of the functional sounds presented.

\subsubsection{Experimental Study Design}
This study adapted the design of Nwagwu et al. \cite{nwagwu2024benefits}, using a Kinova Gen3 manipulator in place of a mobile robot. Participants completed a Lego-building task across three trials, during which the robot provided pieces as needed under three sound conditions: \textit{Consequential}, \textit{Functional}, and \textit{Spatial} (Figure \ref{fig:experimentC_sound_conditions}). For each trial, the robot delivered pieces at a randomly selected time between 15s and 45s after trial onset, preventing participants from anticipating robot movement while allowing them time to engage with the task. Participants were given an additional minute to continue building after the robot’s movement ceased.

After each trial, participants completed a brief survey evaluating the robot and the sounds, without prior information about the sound conditions. Following the three trials, a structured post-experiment survey asked participants to reflect on all conditions, provide opinions and recommendations, and, after replaying the two augmented sound designs with explanations of their implementation, indicate their preferred sound condition and offer additional comments.

\begin{table}[t]
\centering
\small
\caption{Experiment C: Structure of the RoSAS (Robot Social Attributes Scale), consisting of three factors with six items each.}
\label{tab:rosas}
\begin{tabular}{p{3cm} p{10cm}}
\toprule
\textbf{Factor} & \textbf{Items} \\
\midrule
Warmth &
emotional, feeling, happy, organic, compassionate, social \\
Competence &
competent, capable, knowledgeable, interactive, responsive, reliable \\
Discomfort &
scary, strange, awkward, dangerous, aggressive, awful \\
\bottomrule
\end{tabular}
\end{table}
\begin{table}[t]
\centering
\small
\caption{Experiment C: Structured survey questions assessing participant perceptions of the sound conditions. Likert-scale items were rated on a 1--7 scale. All other items were free-response.}
\label{tab:experimentC_questions}
\begin{tabular}{p{15cm}}
\toprule
\textbf{Likert-scale and Open-ended Items} \\
\midrule

\textit{Consequential Sound Condition} \\
\midrule
Which aspects of the robot's natural sounds did you like? \\
Which aspects of the robot's natural sounds did you dislike? \\
Any other thoughts on the consequential sound condition? \\

\midrule
\textit{Functional Sound Conditions (Functional / Spatial)} \\
\midrule
The sounds from the [speaker/AR device] were clear and easy to interpret. (Likert) \\
I could easily distinguish between the different sounds from the [speaker/AR device]. (Likert) \\
I understood what each sound from the [speaker/AR device] was meant to communicate. (Likert) \\
In your opinion, what were the primary functions of the [non-spatial/spatial] sounds? \\
The sounds from the [speaker/AR device] were pleasant to listen to. (Likert) \\
Which aspects of the [non-spatial/spatial] sounds did you like? \\
Which aspects of the [non-spatial/spatial] sounds did you dislike? \\
If you could change the [non-spatial/spatial] sounds, what changes would you make? \\
Any other thoughts on the [non-spatial/spatial] sounds? \\

\midrule
\textit{General Spatial Sound Utility} \\
\midrule
Would spatial sounds help or hinder interactions if the robot were mobile? Why? \\
Would your opinion change if spatial sounds were emitted from eyeglasses? How so? \\

\midrule
\textit{Preference} \\
\midrule
Which condition did you prefer? Please explain your choice. \\

\bottomrule
\end{tabular}
\end{table}

\subsubsection{Measures}
The post-trial survey was the \gls{rosas}, an 18-item Likert-scale questionnaire that assesses human perceptions of robots across three dimensions: warmth, competence, and discomfort \cite{rosas} (Table \ref{tab:rosas}). The post-experiment questionnaire was a structured survey developed by the authors, drawing inspiration from prior studies, that combined Likert-scale and open-ended questions. Its purpose was to capture participants’ evaluations of each sound condition, their understanding of the functional sound design, and their perspectives on the broader role of sound in \gls{hri}. The questions from the structured survey are shown in Table \ref{tab:experimentC_questions}. 

\subsubsection{Analysis}
Responses from the post-trial survey were analyzed using a repeated-measures analysis of variance (ANOVA) with a significance threshold of $\alpha = 0.05$. The assumption of normality was evaluated using the Shapiro–Wilk test, and the Greenhouse–Geisser correction was applied when violations of sphericity were detected. Post hoc comparisons were performed using paired-samples t-tests. When the normality assumption was not met, a Friedman test was conducted as a nonparametric alternative, followed by post hoc Wilcoxon signed-rank tests. Qualitative responses from the post-experiment survey were analyzed using thematic coding.

\subsection{Rationale for Independent Experiments}
Experiments A, B, and C were deliberately designed to be independent, ensuring that outcomes from any one experiment would not influence the others. This approach was motivated by both practical considerations, such as time constraints, and the distinct objectives of each experiment. Specifically, Experiment A examines the perceptual effects of consequential sounds, Experiment B evaluates spatial localization accuracy in \gls{aar}, and Experiment C investigates the effects of both \gls{aar}-based and non-\gls{aar}-based sound designs in a collaborative task. Designing the experiments independently allowed us to tailor procedures and measurements to each specific goal while minimizing potential carryover or ordering effects, providing rationale for maintaining a fixed experiment order for all participants. 

Although Experiments B and C were conducted independently, we recognize that a participant’s ability to localize augmented spatial sounds in Experiment B may be related to their sound preferences (non-\gls{aar} vs \gls{aar}) in Experiment C. We address and explore this potential relationship in Section 5. 

\section{Results}
\subsection{Participants}

\begin{figure}
    \centering
    \includegraphics[width=1.0\linewidth]{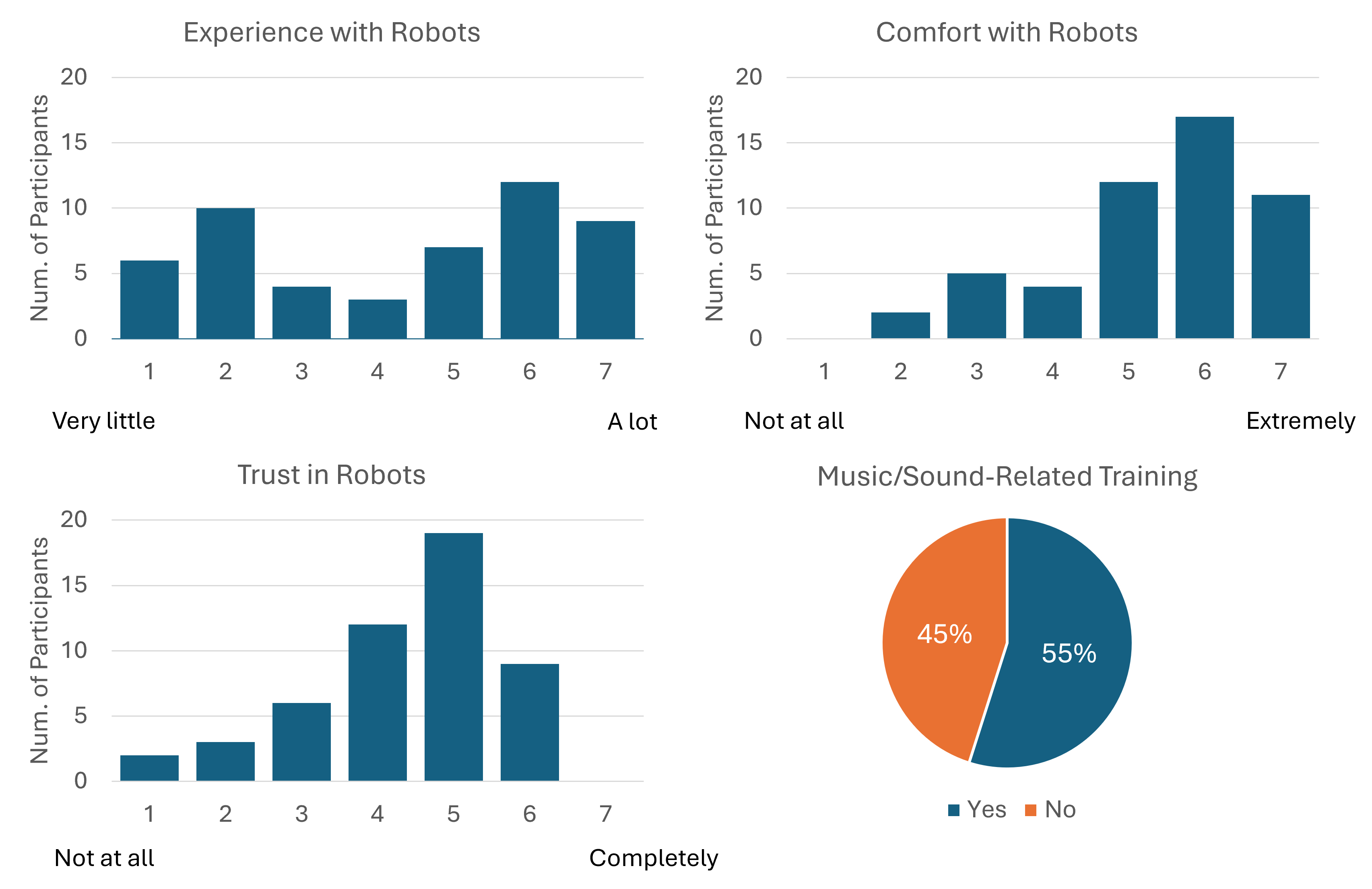}
    \caption{Participant demographics (N=51). Information regarding participants’ music and sound-related training was collected following the completion of the study.}
    \label{fig:demographics}
\end{figure}

A total of 51 participants (30 female, 20 male, 1 preferred not to disclose), aged 19–68 years (M = 26, SD = 8.0), took part in the study. Participants self-reported their experience, comfort, and trust in robots using a 7-point scale. Of the sample, 39.2\% reported low experience with robots (ratings $\leq$ 3), while 21.6\% reported low comfort and trust (ratings $\leq$ 3). Additionally, 54.9\% of participants indicated prior experience with music or sound-related activities. Figure \ref{fig:demographics} shows the distribution of responses for the self-reports. 

All participants completed the full experimental protocol. However, due to experimenter error or technical failures of the HoloLens 2, data from three participants were excluded from Experiment A and data from ten participants were excluded from Experiment C. Consequently, the final sample sizes were 48 participants for Experiment A, 51 participants for Experiment B, and 41 participants for Experiment C. As the three experiments were designed and analyzed as independent evaluations (see Section 3.4), retaining all valid data for each experiment maximizes statistical power and preserves the integrity of the experimental design. Accordingly, results are reported using all available data for each experiment rather than constraining analyses to a uniform participant count across experiments.

The following sections summarize the findings from the three experiments. 

\subsection{Experiment A Results}

\subsubsection{Likert Responses.} 
\begin{figure*}[h] 
    \centering
    \includegraphics[width=\textwidth]{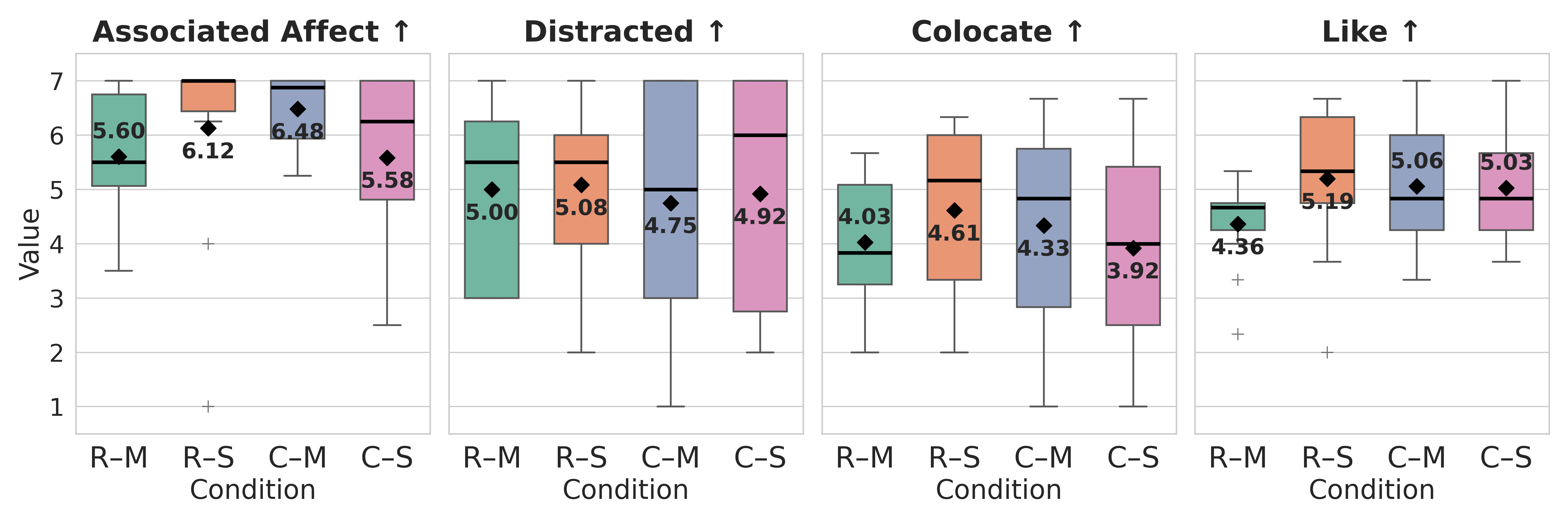}
    \caption{Experiment A: Box-and-whisker plots illustrating participant responses across the four experimental conditions and four perceptual scales in Experiment A. Higher scores indicate more positive perceptions. Black diamonds represent mean values, black horizontal lines indicate medians, plus signs denote outliers, boxes correspond to the interquartile range (25th–75th percentiles), and whiskers extend to 1.5 times the interquartile range. (N = 48)}
    \label{fig:experimentAresults}
\end{figure*}

Figure \ref{fig:experimentAresults} illustrates participants' responses to the Likert questionnaire using box plots, Table \ref{tab:experimentAassumptions} presents the results of normality and homogeneity tests for the four measures, and Table \ref{tab:experiment_A_anova_results} summarizes the primary statistical findings from the two-way ANOVA. Homogeneity of variance was satisfied for all measures (Levene’s tests, all $p > .30$). Residuals deviated from normality for Associated Affect, Distracted, and Colocate (Shapiro-Wilk tests, all $p < .05$), whereas residuals for Like were normally distributed ($p = .60$). To ensure the robustness of our results, we also conducted \gls{art} ANOVAs for these measures, as a non-parametric check, confirming that the observed patterns of main and interaction effects were consistent across methods. We report the results of the two-way ANOVA for clarity and comparability with prior work, as it provides interpretable F-values, p-values, and effect sizes. For factorial designs with balanced sample sizes, the two-way ANOVA is robust to modest deviations from normality, and the \gls{art} ANOVA confirmed the robustness to the violations of normality. Across all measures, there were no statistically significant main effects of Interaction Mode or Sound, nor any significant Interaction Mode $\times$ Sound interactions (all $p > .08$). 

\begin{table}[ht]
\centering
\caption{Experiment A: Assessment of statistical assumptions for all dependent measures. Normality was evaluated using the Shapiro–Wilk test, where significant results indicate deviations from a normal distribution. Homogeneity of variance across conditions was assessed using Levene’s test, with non-significant results indicating that the assumption was satisfied. All tests were evaluated at a significance level of $\alpha = 0.05$.}
\label{tab:experimentAassumptions}
\begin{tabular}{l l l l}
\hline
Measure & Significance Level ($\alpha$) & Normality (Shapiro–Wilk) & Homogeneity (Levene) \\
\hline
Associated Affect & 0.05 & Deviated (W=0.847, p<0.001) & Satisfied (F=0.888, p=0.455) \\
Distracted & 0.05 & Deviated (W=0.892, p=0.0003) & Satisfied (F=1.253, p=0.302) \\
Colocate & 0.05 & Deviated (W=0.950, p=0.039) & Satisfied (F=0.918, p=0.440) \\
Like & 0.05 & Normal (W=0.981, p=0.602) & Satisfied (F=1.100, p=0.359) \\
\hline
\end{tabular}
\end{table}

\begin{table}[ht]
\centering
\caption{Experiment A: Two-way ANOVA results for all measures (Interaction Mode × Sound).}
\label{tab:experiment_A_anova_results}
\begin{tabular}{l l l l l l}
\hline
Measure & Effect & F & df & $\eta^2$ & p-value \\
\hline
Associated Affect & Interaction Mode & 0.171 & 1, 44 & 0.004 & 0.681 \\
& Sound & 0.217 & 1, 44 & 0.005 & 0.644 \\
& Interaction × Sound & 3.091 & 1, 44 & 0.065 & 0.086 \\
Distracted & Interaction Mode & 0.136 & 1, 44 & 0.003 & 0.714 \\
& Sound & 0.049 & 1, 44 & 0.001 & 0.826 \\
& Interaction × Sound & 0.005 & 1, 44 & 0.000 & 0.942 \\
Colocate & Interaction Mode & 0.166 & 1, 44 & 0.004 & 0.686 \\
& Sound & 0.030 & 1, 44 & 0.001 & 0.862 \\
& Interaction × Sound & 1.095 & 1, 44 & 0.024 & 0.301 \\
Like & Interaction Mode & 0.667 & 1, 44 & 0.014 & 0.419 \\
& Sound & 1.553 & 1, 44 & 0.032 & 0.219 \\
& Interaction × Sound & 1.775 & 1, 44 & 0.037 & 0.190 \\
\hline
\end{tabular}
\end{table}

\subsubsection{Responses to Post-experiment Open-ended Sound-Related Questions.} Responses from participants in the sound conditions (\textit{Recorded} and \textit{Collocated}, N=24) regarding the sounds produced by the Kinova Gen3 were organized into three overarching themes: volume and intensity of the produced sounds, desired sounds for this particular robot, and potential for functional sounds (Table \ref{tab:experimentA_robot_sound_thematic_coding}).

\textbf{Volume / Intensity}: Most participants in both groups reported that the Kinova's sounds were at an acceptable volume (C: 4; R: 5). However, several participants also indicated that some sounds were too loud, particularly in the collocated group (C: 4; R: 2), and a smaller number did not notice the sounds at all (C: 1; R: 2). This variability in responses indicate that while the Kinova’s sounds were generally tolerable, proximity increased sensitivity to its mechanical noises.

\textbf{Desired Sound Type}: Participants expressed subjective preferences for the sound profile they considered most appropriate and desirable for the Kinova robot, informed by their exposure to the produced sounds. Silent or very quiet sounds were the most commonly mentioned (C: 4; R: 6), followed by natural, soothing, or whimsical sounds (C: 3; R: 4) and white noise or low hums (C: 2; R: 2). Mechanical sounds such as motor noise were noted less frequently (C: 1; R: 1), and only one participant in each group highlighted the importance of environmental fit. These findings indicate that participants generally prefer robotic manipulators to produce unobtrusive, calming sounds, with some attention to recognition of robot movement.

\textbf{Functional Design}: Participants also identified potential for functional sounds in the Kinova's sound profile. Auditory warnings or signals to indicate robot actions were mentioned by several participants (C: 2; R: 3), while conveying emotion or character through sound was rarely noted (C: 1; R: 0). The reference to functional sound indicates that some participants value auditory feedback primarily for informational or safety-related purposes, especially when the robot is directly interacting with a human.

Overall, the thematic analysis highlights that volume control and unobtrusive sound design are key considerations for robot sound, while functional auditory cues support user awareness of robot actions.

\begin{table}[h!]
\centering
\small
\caption{Experiment A: Thematic coding of participant responses regarding robot sounds for participants in the \textit{Sound} conditions (N=24). Frequencies are reported separately for the \textit{Collocated} (C) and \textit{Recorded} (R) groups.}
\label{tab:experimentA_robot_sound_thematic_coding}
\begin{tabular}{p{3cm} p{2cm} p{7cm} c}
\hline
\textbf{Theme} & \textbf{Facet} & \textbf{Representative Quotes} & \textbf{Frequency (C / R)} \\
\hline
\multirow{3}{*}{\parbox{3cm}{Volume / Intensity}} 
& Acceptable Volume & ``Good – do not affect me negatively'' / ``sounds good and wasn’t too loud or annoying'' & 4 / 5 \\
& Too Loud & ``a little too loud if I was going to use it in my home…'' / ``Pretty loud'' & 4 / 2 \\
& Unnoticed & ``I did not notice any sounds.'' / ``I did not notice them a lot'' & 1 / 2 \\
\hline
\multirow{5}{*}{\parbox{3cm}{Desired Sound Type}} 
& Silent & ``silent'' / ``As quiet as possible'' & 4 / 6 \\
& Natural / Soothing / Whimsical & ``Maybe smoother and more natural'' / ``maybe softer/more continuous than it does now'' & 3 / 4 \\
& White Noise & ``low hum, like white noise that makes me fall asleep'' / ``like white noise or any gentle whirring that can fade into the background'' & 2 / 2 \\
& Mechanical & ``The sound of moving motors, so that I know it is moving...'' / ``I like the sound of stepper motors slowly accelerating'' & 1 / 1 \\
& Environmental & ``I would try to make it fit in better with the sounds in the environment…'' / ``…less noticeable, blends into workspace'' & 1 / 1 \\
\hline
\multirow{2}{*}{\parbox{3cm}{Functional Design}} 
& Warnings & ``Maybe like a beep sound indicating…when it is done with a sequence'' / ``make alert to reminder users when picking up and taking off'' & 2 / 3 \\
& Conveying Emotion & ``…some manufactured/generated sounds for character + comfort, like beeps while processing?''  & 1 / 0 \\
\hline
\end{tabular}
\end{table}

\subsection{Experiment B Results}
\begin{figure*}[h!]  
    \centering
    \includegraphics[width=1.0\textwidth]{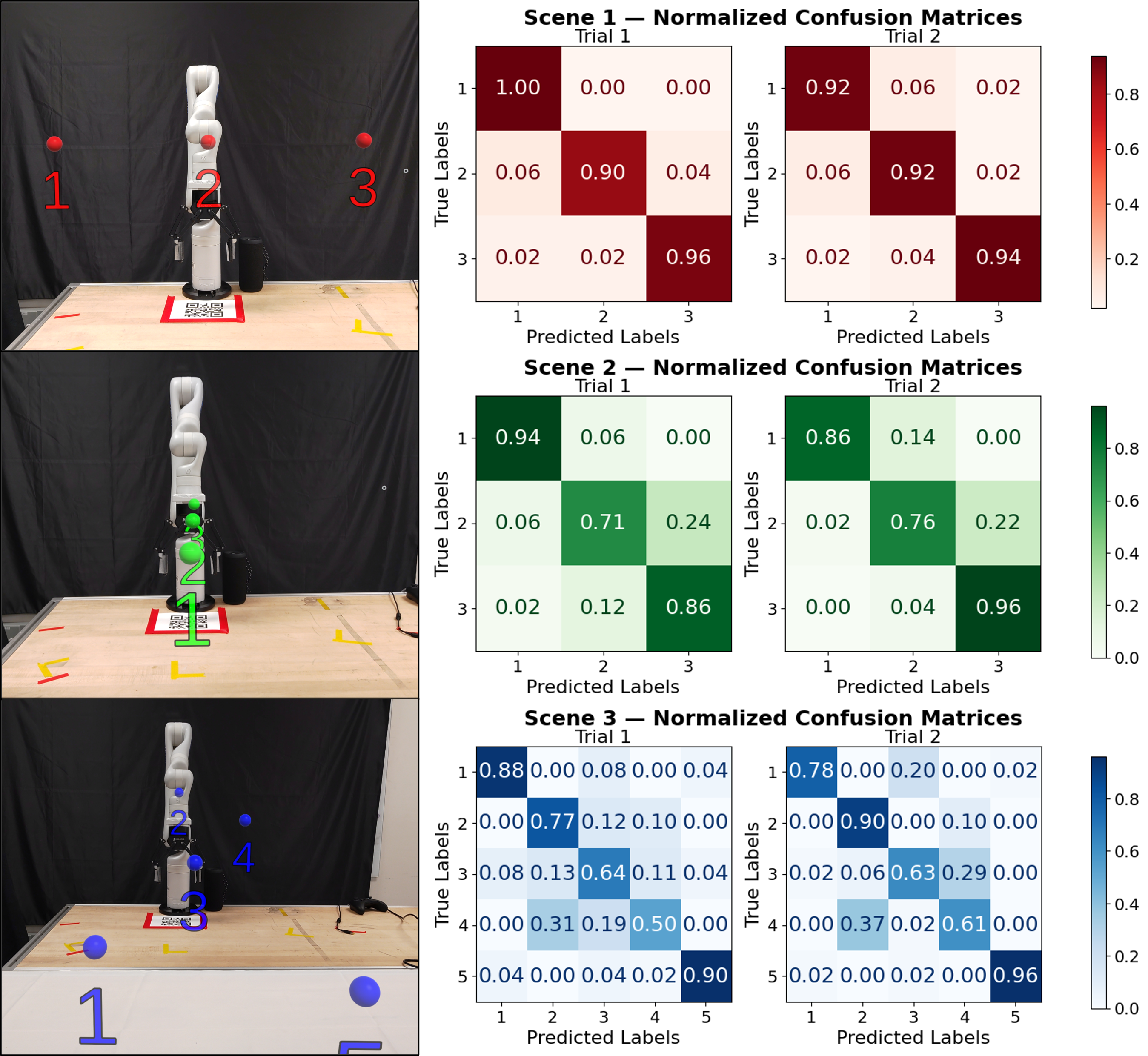}
    \caption{Experiment B: Normalized confusion matrices aggregated across all participants for three scenes, with two trials per scene. Darker shades indicate higher classification accuracy. (N = 51) }
    \label{fig:experimentBresults}
\end{figure*}

Figure~\ref{fig:experimentBresults} presents the three \gls{ar} scenes along with group-level confusion matrices for each trial. Classification accuracy decreased systematically with increasing scene complexity: Scene~1 had the highest performance (95\% and 93\% across trials), followed by moderate performance in Scene~2 (84\% and 86\%), and the lowest accuracy in Scene~3 (74\% and 78\%). Accuracy declined slightly from Trial~1 to Trial~2 in Scene~1, whereas modest improvements were observed across trials in Scenes~2 and~3.

To determine whether these observed differences reflected reliable effects rather than sampling variability, we additionally fit a linear regression model predicting binary correctness from scene and label identity. This analysis was conducted as a confirmatory check and was not intended as a primary reporting metric. The regression results corroborated the patterns evident in the accuracy values and confusion matrices, revealing significant effects of both scene and label identity that aligned with the descriptive findings. Importantly, no regression results contradicted the observed accuracy trends. For clarity and readability, detailed regression coefficients are omitted.

\subsection{Experiment C Results}

\subsubsection{ROSAS Responses}
\begin{figure*}[h!]  
    \centering
    \includegraphics[width=\textwidth]{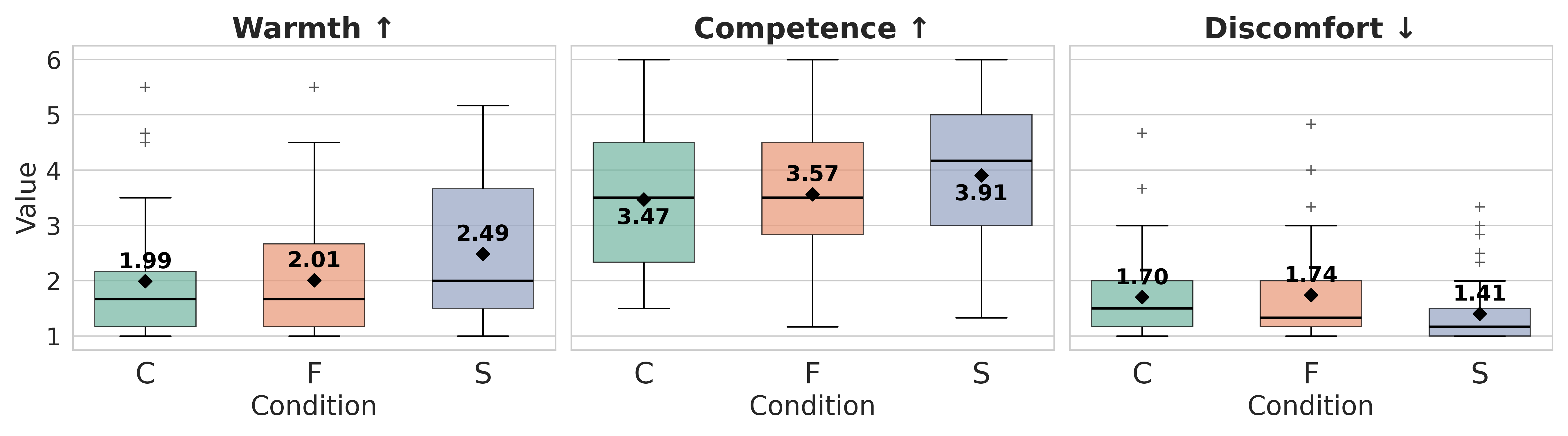}
    \caption{Experiment C: Box-and-whisker plots illustrating participant responses across the three experimental conditions and three attribute scales. Plot elements are structured as described in Figure \ref{fig:experimentAresults}. (N = 41)}
    \label{fig:experimentCresults}
\end{figure*}

Figure \ref{fig:experimentCresults} presents participants' responses as box plots and Table \ref{experimentC_statistics} summarizes the main statistical outcomes. 

\textbf{Warmth:} Shapiro–Wilk tests indicated violations of normality across all conditions (Consequential: $W = 0.812$, $p < .001$; Functional: $W = 0.843$, $p < .001$; Spatial: $W = 0.913$, $p = .004$); therefore, a Friedman test was conducted. The omnibus test revealed no significant differences in perceived warmth across conditions, $\chi^2(2) = 3.17$, $p = .205$, Kendall's $W = 0.039$. Mean warmth ratings were similar across conditions (Consequential: $M = 1.99$; Functional: $M = 2.01$; Spatial: $M = 2.49$). 

\textbf{Competence:} Normality was met for all conditions (Consequential: $W = 0.946$, $p = .052$; Functional: $W = 0.961$, $p = .170$; Spatial: $W = 0.954$, $p = .098$), and sphericity was assumed. A repeated-measures ANOVA indicated no significant effect of condition on perceived competence, $F(2, 80) = 0.97$, $p = .384$, $\eta^2_g = 0.020$. Mean competence ratings were comparable across conditions (Consequential: $M = 3.47$; Functional: $M = 3.57$; Spatial: $M = 3.91$). 

\textbf{Discomfort:} Shapiro–Wilk tests revealed violations of normality for all conditions (Consequential: $W = 0.811$, $p < .001$; Functional: $W = 0.802$, $p < .001$; Spatial: $W = 0.712$, $p < .001$), prompting the use of a Friedman test. No significant differences in discomfort were observed across conditions, $\chi^2(2) = 5.03$, $p = .081$, Kendall's $W = 0.061$. Mean discomfort ratings were low overall (Consequential: $M = 1.70$; Functional: $M = 1.74$; Spatial: $M = 1.41$).

\begin{table}[h!]
\centering
\caption{Experiment C: Summary of main test results for Warmth, Competence, and Discomfort. 
$\chi^2$ denotes the Friedman test statistic, $F$ denotes the repeated-measures ANOVA statistic, 
$\eta^2_g$ denotes generalized eta-squared as a measure of effect size for ANOVA, 
and Kendall's $W$ denotes the effect size for Friedman tests. $p$ represents the probability of observing the data under the null hypothesis.}
\begin{tabular}{lccccc}
\hline
\textbf{Measure} & \textbf{Normality} & \textbf{Test} & \textbf{Statistic} & \textbf{Effect Size} & \textbf{p-value} \\
\hline
Warmth & Non-normal & Friedman & $\chi^2(2) = 3.169$ & Kendall's $W = 0.039$ & 0.2051 \\
Competence & Normal & RM-ANOVA & $F(2.00, 80.00) = 0.970$ & $\eta^2_g = 0.0197$ & 0.3836 \\
Discomfort & Non-normal & Friedman & $\chi^2(2) = 5.027$ & Kendall's $W = 0.061$ & 0.0810 \\
\hline

\end{tabular}
\label{experimentC_statistics}
\end{table}
The post-experiment questionnaire collected participant responses across all three experimental conditions, including open-ended feedback on likes and dislikes for each sound design. For the \textit{Consequential} condition, many participants reported neutral impressions or noted that they barely noticed the sounds, while particularly appreciating that they were quiet and did not distract from the task. For example, Participant 25 stated: ``the sounds by themselves were relatively quiet and didn't bother me." At the same time, several participants described the sounds as overly mechanical, robotic, or unnatural, with some expressing a preference for even quieter cues. Participant 33 commented: ``Makes me feel like it doesn't care that I'm working on something, doesn't care whether or when I'm ready to interact with it or want its help or even notice it." In the \textit{Functional} condition, participants generally appreciated that the sounds clearly reflected the robot’s status and movements, enhancing awareness of its intentions. The high-pitched, musically structured sounds also aligned with expectations from familiar household devices. Participant 28 explained: ``I liked that it was higher pitched and made the robot seem a little more friendly and similar to other things I have in my house that beep like the dishwasher and laundry etc., so it was more familiar and less weird." However, several participants found the sounds to be too loud, harsh, or startling, which sometimes caused stress or distraction from the task. Participant 46 noted: ``I didn't like the strong beeping sounds from the robot. It made me get worried that I wasn't doing the task fast enough." For the \textit{Spatial} condition, participants highlighted the calming and pleasant musical accompaniment, which contributed to a more positive perception of the robot. Some participants also reported that these sounds enhanced their awareness of the robot’s spatial movements. Participant 17 stated: ``They were cute and fun. More pleasant to listen to than the [\textit{Functional}] sounds. [I] like that they alerted me to the robot's movements." Nonetheless, a subset of participants noted that the meaning of the spatial sounds was less immediately interpretable. Participant 23 observed: ``The cues were much harder to interpret than the speaker sound." Finally, a few participants reported that the weight and bulkiness of the \gls{ar} headset were their primary dislikes, independent of the sound conditions.

The following analysis focuses on additional qualitative feedback specifically on the two augmented sound conditions, \textit{Functional} and \textit{Spatial}, which are of primary interest to this study.

\begin{figure}[h!]
    \centering
    \includegraphics[width=1.0\linewidth]{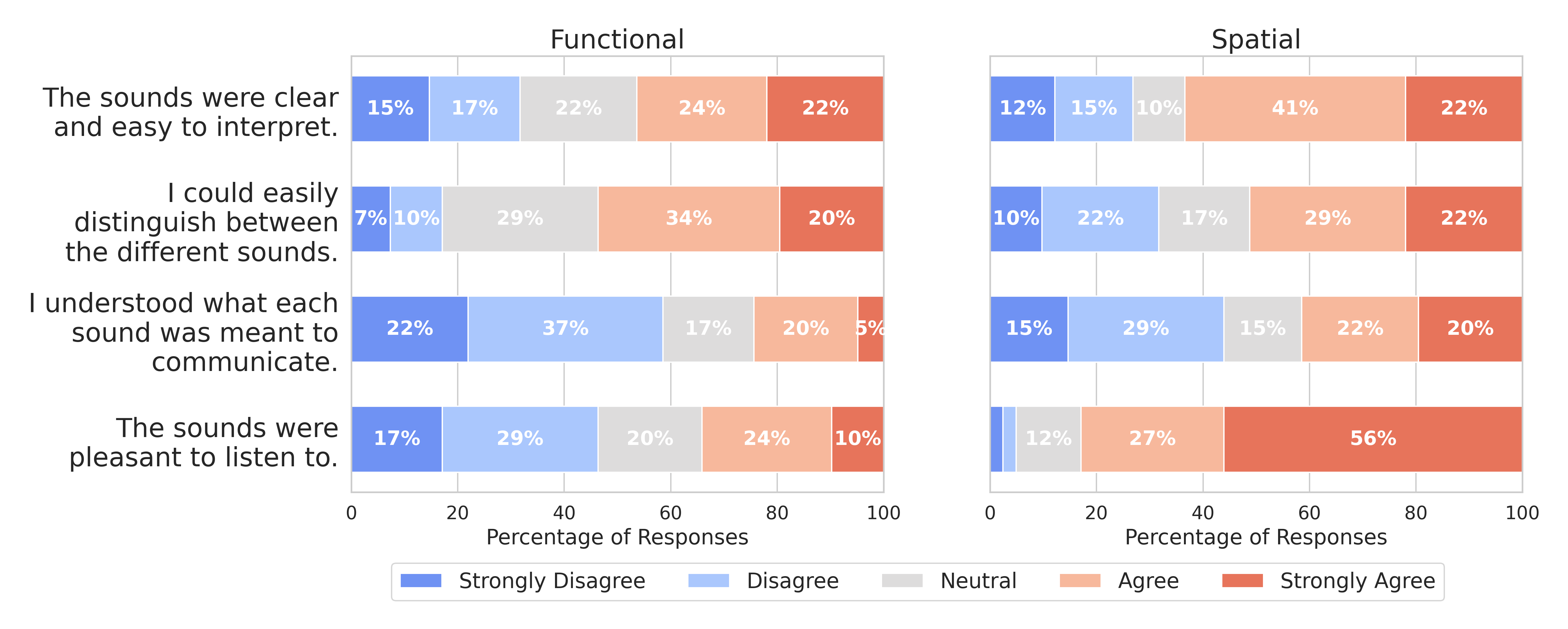}
    \caption{Experiment C: Responses to four Likert-scale questions for the two augmented sound conditions.}
    \label{fig:experimentC_likert_augmented_conditions}
\end{figure}

\textbf{Sound Condition Design Attributes:} Figure \ref{fig:experimentC_likert_augmented_conditions} illustrates the distribution of participant responses to four Likert-scale items assessing perceptions of the sound designs in the two augmented conditions (\textit{Functional} and \textit{Spatial}). For the \textit{Functional} condition, most participants agreed that the sounds were perceptually distinguishable; however, a comparatively smaller proportion reported confidence in interpreting the semantic meaning or communicative intent of the individual sounds. Responses regarding pleasantness exhibited substantial variability, indicating heterogeneous subjective evaluations.

In contrast, responses for the \textit{Spatial} condition showed stronger agreement overall, with a larger proportion of participants indicating that the sounds were interpretable, distinguishable, and communicative. Despite this general trend, response distributions remained relatively broad, suggesting individual differences in perception and preference. Notably, an overwhelming majority of participants rated the \gls{aar} sound design as pleasant, pointing to a consistently positive affective assessment of this condition.

\textbf{Perceived Sound Condition Function:} Thematic coding of post-experiment survey responses regarding the perceived function of the two augmented sound conditions summarized in Table \ref{tab:thematic_coding_sounds} showed that participants primarily associated the \textit{Functional} sounds with communicating robot movement, task progress, and readiness for interaction, whereas the \textit{Spatial} sounds conveyed these functions plus calming effects and more precise directional cues. The six distinct cues in \textit{Functional} mapped to three functions, while the single continuous musical piece in \textit{Spatial} mapped to five functions, highlighting the potential for spatial sounds to serve both functional and transformative roles in \gls{hri}. Overall, 12.2\% of participants preferred the \textit{Consequential} condition, 31.7\% preferred the \textit{Functional} condition, and 56.1\% preferred the \textit{Spatial} condition.

\begin{figure}[h!]
    \centering
    \begin{minipage}[b]{0.48\linewidth}
        \centering
        \includegraphics[width=\linewidth]{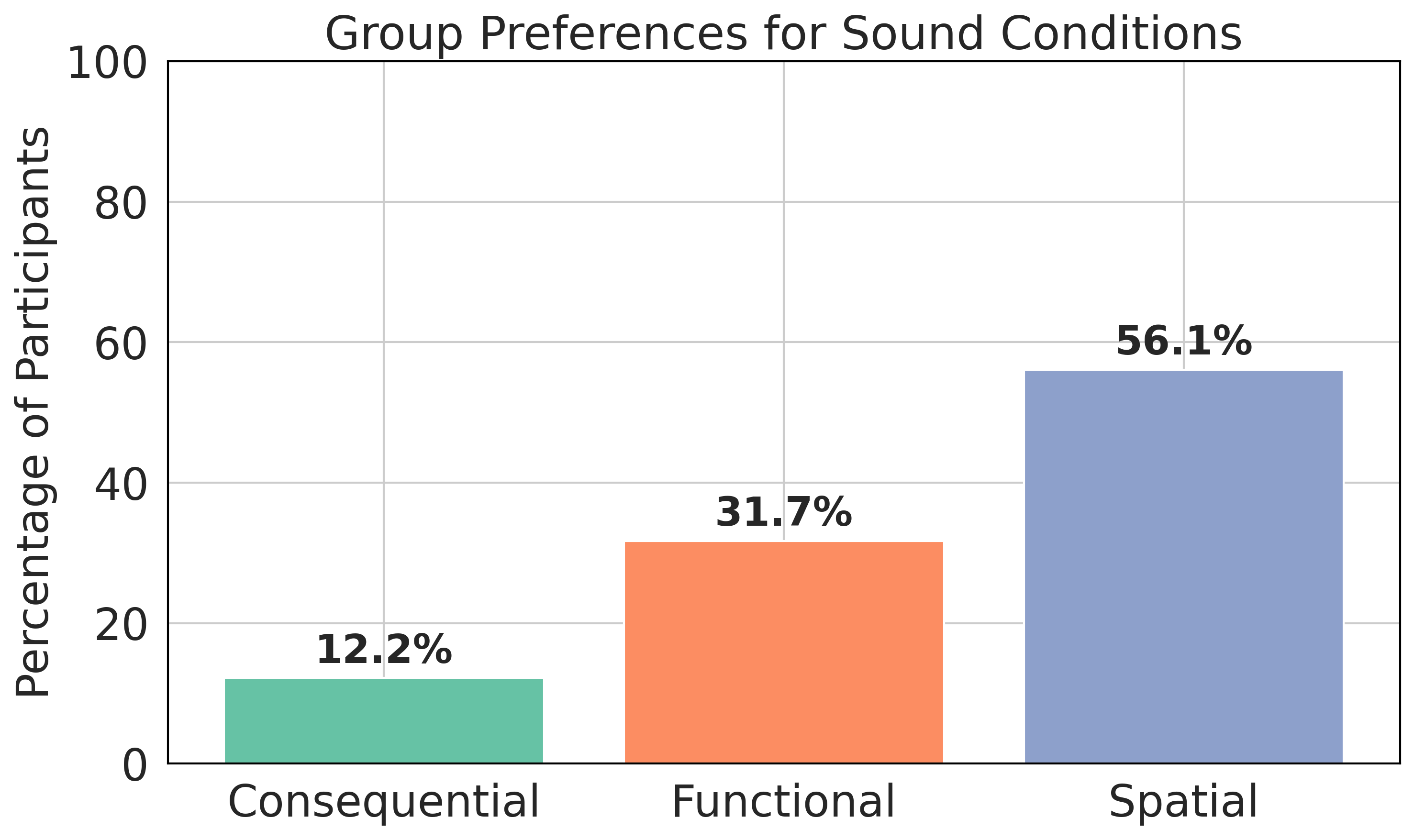}
    \end{minipage}
    \hfill
    \begin{minipage}[b]{0.48\linewidth}
        \centering
        \includegraphics[width=\linewidth]{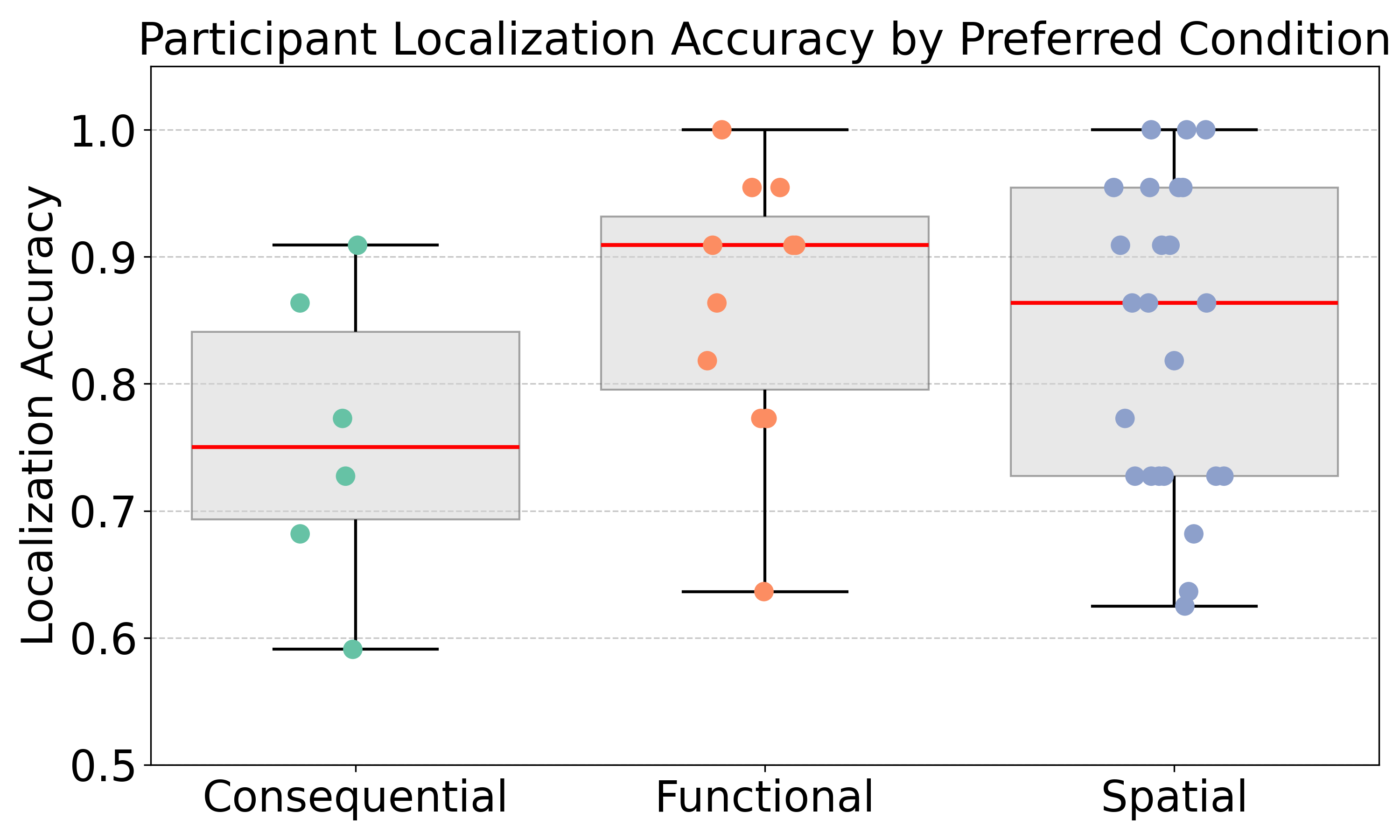}
    \end{minipage}
    \caption{Sound condition preference among all participants (Left) and corresponding participant localization accuracy based on preferred sound condition (Right). (N = 41)}
    \label{fig:combined_sound_accuracy}
\end{figure}

\textbf{Sound Condition Preference:} Figure \ref{fig:combined_sound_accuracy} summarizes participants’ preferences for the three sound conditions following an explanation of the two augmented designs. Participants who favored the \textit{Consequential} sound condition largely indicated that the robot’s natural mechanical sounds provided sufficient information, such as motor sounds during movement and silence when the robot stopped. For example, Participant 18 stated: ``I don't think that human should be keep being informed about robot's movement, and it is sufficient to be notified when human has to interact with robot (e.g. when getting package handed over from the robot). So the consequential sounds, with only natural sound of robot movement, was enough to notify the necessary information. (e.g. when the robot became silent, it means it has stopped, potentially indicating that it's time to get package from the robot)." Those who preferred the \textit{Functional} condition emphasized its clarity and informativeness, noting that the sounds conveyed their intended meanings without unnecessary complexity. Participant 27 commented: ``I think [the \textit{Functional} condition] offers the best option for audio feedback to support user interactions. [The \textit{Spatial} condition] is so musical/relaxing that it was difficult to interpret the changing directions, and [the \textit{Consequential} condition] does not add any audio information." However, the majority of participants preferred the \textit{Spatial} condition. Nearly all cited the comforting nature of this design, which reduced anxiety during task completion and robot interaction. Participants also appreciated the richness of information about the robot’s movements, although some only fully understood this after receiving the researcher’s explanation. Participant 31 explained: ``It conveyed a lot more information. Since the sounds were changing as the robot was moving, I could tell when the arm was going to move and approximately where in what direction the arm was going to go to. The noises were more pleasant to listen to."

\begin{table}
\small
\centering
\caption{Experiment C: Summary of thematic coding for participants’ predicted functions of the two augmented sound conditions (N = 41).}
\begin{tabularx}{\linewidth}{p{1.6cm} p{3.2cm} X p{1.8cm}}
\toprule
\textbf{Condition} & \textbf{Predicted Function} & \textbf{Representative Quotes} & \textbf{Frequency} \\
\midrule

\multirow{5}{*}{\textbf{Functional}} 
 & Alerting to Movement & ``To alert me when the robot was moving.'' & 24 \\
 & Indicating Task Progress & ``To inform what stage [the robot] was doing — going to pick up object, object ready for pickup, dropping the bag.'' & 18 \\
 & Requesting Attention / Readiness & ``Human should be prepared to get the package.'' & 12 \\
 & Not Noticed / Unclear & ``I honestly didn’t realize there were sounds coming from the speaker.'' & 5 \\
 & Communicating Success or Error & ``The medium-pitch sound conveyed [the robot] didn’t find what it was looking for; the higher one said ‘I’ve found it now!’'' & 1 \\

\midrule

\multirow{7}{*}{\textbf{Spatial}} 
 & Alerting to Movement & ``The sounds alerted me to when the robot was starting or moving.'' & 16 \\
 & Calming Effect & ``[The sounds were] trying to make me feel calmer and more relaxed with the robot.'' & 15 \\
 & Directional / Locational Cues & ``[The sounds] moved around the head according to the robot to indicate its position.'' & 15 \\
 & Requesting Attention / Readiness & ``The music is getting dramatic—get ready to take the bag.'' & 10 \\
 & Indicating Task Progress & ``I assume the spatial sounds were trying to communicate the stage of the task — picking up the bag, moving into position.'' & 8 \\
 & Immersion & ``Immersion, cut out outside noise to focus on the task.'' & 3 \\
 & Not Noticed / Unclear & ``I could not distinguish the different sounds going on at once.'' & 3 \\

\bottomrule
\end{tabularx}
\label{tab:thematic_coding_sounds}
\end{table}

\section{Discussion}
\subsection{Key Findings} 
\subsubsection{Experiment A} \textbf{H1} and \textbf{H2} were not supported. Responses to the Likert-scale measures indicated no significant differences across groups in affective reactions, liking, or preference for collocation. Qualitative feedback from participants in the \textit{Sound} conditions consistently described the Kinova Gen3 robot as producing minimal consequential noise, with the sounds generally perceived as tolerable even when participants were physically close to the robot. These findings contrast with prior results reported in \cite{allen}, further underscoring the platform-dependent nature of consequential sound perception. This divergence highlights the need for caution when generalizing sound-related effects across robotic systems and suggests that additional investigations involving a broader range of platforms are necessary. Notably, participants’ qualitative descriptions of their desired robot sound profiles align with existing literature, reinforcing prior observations that users tend to prefer robots that are quiet or produce naturalistic sounds \cite{wang, allen_hri, zhang_kawaii}.

\subsubsection{Experiment B} \textbf{H3} was supported. Participants demonstrated high accuracy in localizing sounds within the \gls{aar} environment. As scene complexity increased, localization accuracy for lateral sound cues remained consistently high, whereas accuracy declined for sounds originating directly in front of participants. This pattern aligns with established findings in human auditory perception, where localization performance is shaped by factors such as source distance, head orientation, reverberation, interaural time differences, and interaural phase differences \cite{carlini, kolarik2016auditory}. These results carry important implications for the design of \gls{aar} interfaces, highlighting the need for careful deployment of spatial audio cues. Specifically, they suggest that the effectiveness of spatial sounds may vary systematically with source positioning and inherent perceptual constraints.

\subsubsection{Experiment C} \textbf{H4} and \textbf{H5} were not supported, as no statistically significant differences were observed across the three sound conditions for perceived robot warmth, competence, or human discomfort. Qualitative feedback, however, revealed substantially stronger and more nuanced participant opinions. Although many participants appreciated the quiet nature of the consequential sound condition, the majority expressed a preference for sound designs that conveyed functional information. This observation reinforces insights from Experiment A, where several participants indicated a desire for informative auditory cues. Notably, participants diverged in their reactions to how such cues were implemented. Some favored the directness of the \textit{Functional} condition, whereas others described these sounds as uncomfortable, citing their abrupt, mechanical, or distracting qualities—perceptions consistent with prior findings \cite{nwagwu2024benefits}. In contrast, most participants responded positively to the \gls{aar}-based \textit{Spatial} sound design. These sounds were frequently described as pleasant and were associated with perceptions of the robot as more approachable, while also contributing to feelings of calmness and improved task focus. This response aligns with prior work suggesting that naturalistic or rhythmic auditory feedback can positively shape human perceptions of robotic systems \cite{allen_hri}. Importantly, this condition also yielded a key finding of relevance to the \gls{hri} community. Beyond improving subjective experience, transformative sound designs delivered through spatial audio and \gls{aar} effectively communicated functional information, as reflected in the support of \textbf{H6}. The thematically coded qualitative responses indicated that participants were largely able to infer the intended functions of both the \textit{Functional} and \textit{Spatial} sound conditions without prior explanation or explicit notification that sounds would be presented, despite reported uncertainty regarding their meanings.

Collectively, these findings highlight both the communicative potential of well-designed auditory cues in \gls{hri} and the broader participant preference for functionally informative robot sounds, thereby contributing to the relatively limited body of work on functional sound in robotics.

\subsection{Trends between Spatial Sound Localization Performance and Sound Condition Preference}

Among participants with valid data in both Experiments B and C, we analyzed whether individual sound localization accuracy was associated with participants’ preferred sound conditions. Individual localization accuracy was computed as the proportion of spatial audio cues correctly identified across all three scenes in Experiment B, defined as the number of correctly identified cues divided by the total number of cues. Figure \ref{fig:combined_sound_accuracy} presents a scatter plot of localization accuracy for each preferred sound condition. While participants who preferred the \textit{Functional} condition exhibited the highest average localization accuracy, variability within each group indicates that there is no clear correlation between higher localization performance and sound preference.

\subsection{Implications for Sound Design in Robotics}
\begin{table}[t]
\centering
\small
\caption{Survey item asking participants to rank the importance of different sound functions. Participants ranked each function from most important to least important.}
\label{tab:sound_ranking_questions}
\begin{tabular}{p{12.5cm}}
\toprule

Please rank the following functions for communicating through sound from most important to least important: \\
\midrule
Safety \\
Robot status (on/off/idle) \\
Task information \\
Information regarding the robot's movement in space \\

\bottomrule
\end{tabular}
\end{table}

 \begin{figure*}
    \centering
    \includegraphics[width=\textwidth]{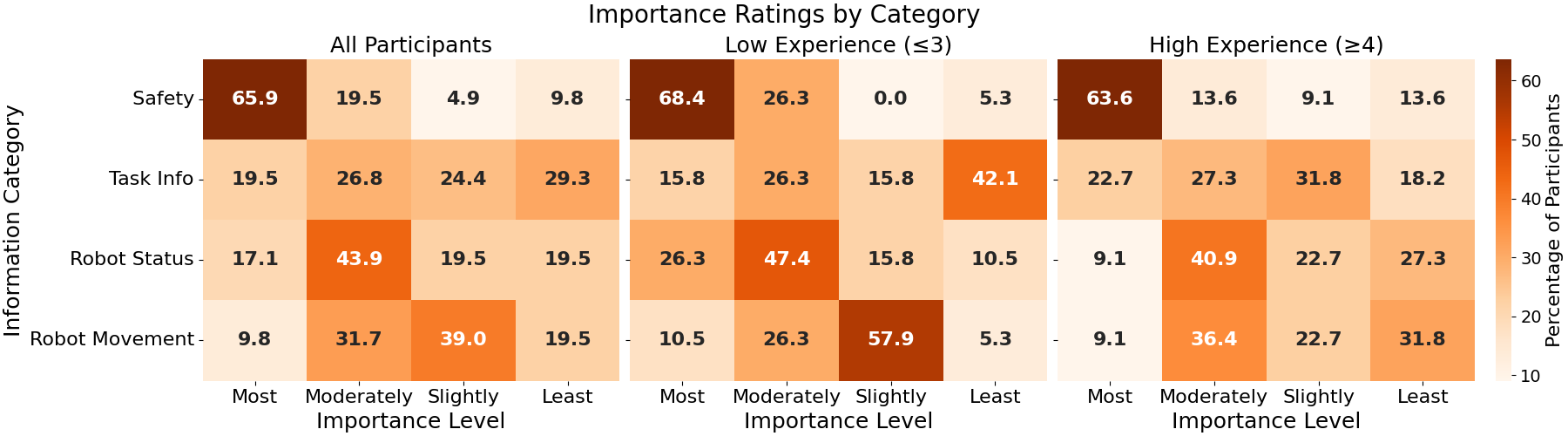}
    \caption{Heatmaps depicting participants’ ratings of the importance of communicating four distinct categories of information through sound. The left panel shows overall ratings across all participants (N = 41), the middle panel shows ratings from participants reporting low experience with robots (N = 19), and the right panel shows ratings from participants reporting high experience with robots (N = 22).}
    \label{fig:soundcategoryheatmap}
\end{figure*}

The results of this study emphasize the importance of integrating functional audio into \gls{hri} while minimizing consequential noise. Prior work has similarly highlighted the value of quiet robot design; however, our findings reveal that even relatively quiet platforms, such as the Kinova Gen3, could benefit from further reductions in sound, as many participants indicated a preference for an even quieter experience. These observations underscore the need for careful attention to the volume and intensity of robot-generated sounds. Ignoring this design consideration may result in robots that are less suitable for homes, workplaces, and public spaces, potentially hindering user adoption.

We propose that the design space for functional sounds in \gls{hri} is relatively large, particularly when leveraging transformative sounds. As with other types of interface design, the context of interaction is critical. Key design considerations include the directness of the sounds—whether they should be explicit or more implicit, as exemplified by the \textit{Spatial} sound condition; the volume of the sounds—whether it remains constant or varies throughout the interaction; and the source of the sounds—whether they emanate from a nearby speaker, the robot itself, or a headset worn by the user. Additionally, designers should consider the number of distinct sounds included, as excessive sounds may overwhelm users, as well as whether the sounds should be continuous or intermittent.

Furthermore, designers should carefully consider the specific types of information that should be conveyed to users during interaction. While these information needs are critical, they remain an underexplored aspect of auditory interface design. To investigate this, participants were asked to rank the importance of various categories of information that could be communicated through sound (Table \ref{tab:sound_ranking_questions}). As illustrated in Figure \ref{fig:soundcategoryheatmap}, safety-related information emerged as the highest priority across all participants. We also examined whether these preferences differed based on participants’ prior experience with robots—for example, comparing individuals who regularly interact with robots (e.g., students and researchers) to those who primarily coexist with them. Results showed that experienced users held more divided opinions regarding the importance of task-related information, whereas information about robot status and movement was generally considered less critical by experienced participants compared to those with limited interaction experience.

Sensitivity to auditory stimuli represents another critical factor in the design of functional cues. Several participants reported that their individual sensitivities influenced their perceptions of the sounds. For example, some found the functional sounds excessively loud, while others perceived them as too quiet. Similarly, opinions diverged regarding pitch—some participants described the higher-pitched sounds as unpleasant, whereas others perceived them as contributing to a friendlier robot character. These findings highlight the variability in individual auditory preferences and suggest that personal differences in sound sensitivity, musical taste, and perceptual thresholds may meaningfully influence user experience and the effectiveness of auditory communication in \gls{hri}.

Spatial audio introduces a novel design space in human–robot interaction, particularly when delivered via \gls{aar} on head-mounted \gls{ar} devices. This approach enables auditory information to be communicated at the individual level, as audio can be programmed to exist within specific spatial radii or be restricted to headset users. Such designs allow for localized, personalized sounds that are minimally distracting to others in the surrounding environment. Building on these capabilities, we recommend that designers leverage spatial and functional audio more broadly to enhance communication between robots and humans. Carefully designed auditory cues can improve task-related information transfer, support positive user experiences, and foster greater comfort and trust during collaborative interactions.

\subsection{\gls{aar} Interfaces in the Real World}
As discussed previously, \gls{aar} interfaces designed for \gls{hmd} present substantial opportunities for interaction design. A primary limitation, however, lies in the reliance on headsets that are often perceived as heavy, bulky, and physically uncomfortable. Improving the accessibility of \gls{aar} and auditory interfaces will therefore require advances toward smaller, lighter, and more ergonomic \gls{ar} devices.

In the post-experiment questionnaire, participants were asked whether their perception of the utility of \gls{aar}-based sound design would change if spatial audio were delivered through a pair of eyeglasses rather than the HoloLens 2. Several participants indicated that eyeglasses would be less intrusive and more comfortable, with some noting that such a form factor would increase practicality and likelihood of everyday use. At the same time, participants acknowledged potential technical constraints associated with audio delivery through glasses. One participant articulated this tradeoff, stating, ``I feel that there would be more utility but less impact. Eyeglasses are less intrusive and more natural to wear and interact with than a full-fledged headset. However, there would be serious limitations in the volume of sound, range of frequencies, etc., that might reduce impact and immersiveness.” Other participants suggested that form factor alone would not substantially alter their perceptions, with some expressing skepticism regarding the benefits of spatial audio more broadly. For example, one participant remarked, ``Not really, I’m skeptical of the added benefit of spatial sound. But it is a neat feature. Having it on glasses might make it more useful, since one could learn the ‘calibration’ of where the sound is coming from.”

Additionally, participants were asked how their perceptions of spatial sound might change when such cues are associated with a mobile robot, such as the TidyBot \cite{tidybot}, operating within a home or workplace environment. The majority of participants indicated that spatial sounds would likely enhance interactions. For example, one participant stated, ``I think it would help interactions with the robot. With the spatial sounds, I feel like you can convey information on location, activities, or movements that the robot is doing. Especially in a social setting, I think these noises could make itself known." Participants who anticipated that spatial sounds might hinder interactions frequently cited a preference for quiet environments, particularly in work contexts. As one participant explained, ``[These sounds would] hinder [interactions]- I am someone that likes quiet when I am trying to work, and if I want to listen to sounds, I normally choose the type (music, rain, etc)."

Finally, it is important to acknowledge that real-world environments are already saturated with auditory stimuli, raising legitimate concerns about the potential for additional “sound pollution.” Participants themselves recognized this tension between the benefits and drawbacks of added robot-generated audio. One participant captured this tradeoff, stating, ``I think it could help in that it would make the actions and interactions of the robot create more noise that make it seem more naturalistic or potentially intuitive. Spatial sounds might also make the robot easier to identify and therefore less scary/unknown. However, I could see it hindering interactions with the robot if it is moved into a particular noisy or crowded space, where having spatial sounds could be distracting/annoying/difficult to hear and interpret." This perspective highlights an important design challenge: while spatial and functional sounds may enhance interpretability and user comfort, their effectiveness is inherently context-dependent. Designers must therefore carefully consider environmental noise, user attention, and the risk of auditory overload when integrating sound into human–robot interactions.

Although many open questions remain regarding the practical utility of \gls{aar} in real-world contexts, this study offers evidence of its potential to support information communication and positively influence human perceptions of robots. These findings suggest that spatial sound, when thoughtfully designed, may serve as a valuable interaction modality rather than a purely aesthetic feature. We therefore encourage designers and researchers to build upon this work and further explore how spatial audio can enhance human–robot interactions in real-world environments.

\subsection{Limitations and Future Work}
Experiment A used a between-subjects design with 12 participants per condition, and a larger sample may have revealed stronger perceptual differences. The brief pick-and-place task limited exposure to robot movements, potentially reducing the impact of consequential sounds. In video-based conditions, standardized audio may have affected participants differently depending on auditory sensitivity, and although noise-canceling headphones attenuated most ambient sounds, faint robot noises may still have been audible in the in-person setting. In Experiment B, sound sources were not randomly selected but chosen to test specific scenarios (e.g., lateral vs. medial); while a broader range might have produced different outcomes, observed trends aligned with prior research. Experiment C may have been influenced by the relatively small participant sample. This study was conducted using a stationary manipulator, which demonstrates that spatial audio can be effective even in confined workspaces. However, spatial audio has the potential to be even more beneficial in larger workspaces or when applied to mobile robots. When asked whether spatial sounds (beyond the specific design used in this study) would be advantageous for mobile robots, participants overwhelmingly agreed, citing examples such as improved ability to anticipate and avoid the trajectories of a mobile robot and enhanced sound localizability, particularly given the low consequential audio produced by the robot used in this study. As one participant noted, ``The headset might have spatial stereo capabilities, but that is more helpful in my mind when we are in the center of the experience. For something in front of me, not around, perception and processing of the sounds in different directions is less functional.” These findings suggest that future experiments should investigate the deployment of spatial audio on robots that move around humans to fully evaluate its functional and perceptual benefits.

While this paper explored two specific sound designs, further research is needed to investigate how functional and spatial sounds can be designed beyond discrete chimes or continuous musical pieces to better accommodate the diverse—and sometimes conflicting—sound preferences of users. As noted in prior work, sound design becomes increasingly complex when multiple simultaneous sources are present. Future studies should therefore examine how augmented sound designs perform in multi-robot contexts. Additionally, it will be important to explore how the effectiveness and perception of such sounds may change with age and varying hearing abilities.

\section{Conclusion}

As robots become increasingly integrated into daily life, understanding how sound influences human–robot interaction is essential. This study examined the Kinova Gen3 manipulator to investigate how consequential, functional, and spatial sounds shape human perception and behavior. First, we assessed the impact of consequential sounds across both in-person and video-based settings. Next, we evaluated spatial sound delivered through \gls{aar} in a localization task, providing evidence for the types of spatial auditory cues most effective in potential \gls{aar} and \gls{hri} applications. Finally, we compared three sound conditions—consequential sounds and two augmented functional sound designs (one of which employs \gls{aar})—during a collaborative handover task.

Results indicated that consequential sounds did not negatively affect participants’ perceptions due to the robot’s inherently quiet design. Participants localized lateral sounds with higher accuracy than frontal sounds, consistent with prior findings from studies that did not employ \gls{vr} or \gls{ar} \gls{hmd} technologies. Importantly, spatial sounds delivered via \gls{aar} enhanced perceptions of warmth and reduced discomfort, as revealed through participant qualitative feedback. Participants also indicated that safety is the most critical information to communicate through functional sounds, whether delivered via \gls{aar} or non-\gls{aar} methods, while opinions were more divided regarding the importance of other information, such as robot movement, status, and task-related cues.

Collectively, these findings extend prior work on the perceptual effects of consequential sound, demonstrate the potential of \gls{aar} in \gls{hri}, provide practical considerations for designing functional sounds, and highlight the capacity of spatial audio to serve both functional and transformative roles, advancing the development of intuitive, sound-enhanced human–robot interactions.

\printbibliography










\end{document}